\documentclass[10pt]{article} 
\usepackage[preprint]{tmlr-template/tmlr}

\usepackage{amsmath,amsfonts,bm}

\def\eqref#1{equation~\ref{#1}}

\def\1{\bm{1}}

\def\vtheta{{\bm{\theta}}}
\def\vphi{{\bm{\phi}}}
\def\va{{\bm{a}}}

\def\vp{{\bm{p}}}
\def\vq{{\bm{q}}}

\def\vu{{\bm{u}}}

\def\vx{{\bm{x}}}
\def\vy{{\bm{y}}}

\DeclareMathAlphabet{\mathsfit}{\encodingdefault}{\sfdefault}{m}{sl}
\SetMathAlphabet{\mathsfit}{bold}{\encodingdefault}{\sfdefault}{bx}{n}

\def\sB{{\mathbb{B}}}

\def\sD{{\mathbb{D}}}

\def\sR{{\mathbb{R}}}

\usepackage{hyperref}
\usepackage{url}
\usepackage{graphicx}
\usepackage{float}
\usepackage{amsmath}
\usepackage{amssymb}
\hypersetup{hidelinks}

\usepackage{bm}
\usepackage{tabularx}

\def\vTheta{{\bm{\Theta}}}
\def\vPhi{{\bm{\Phi}}}
\newcommand{\lossidx}[1]{\if\relax\detokenize{#1}\relax\else^{#1}\fi}
\newcommand{\dambiguity}[1][]{\mathcal{A}\lossidx{#1}_{\mathrm{R}}}
\newcommand{\pambiguity}[1][]{\mathcal{A}\lossidx{#1}_{\mathrm{E}}}
\newcommand{\pperformance}[1][]{\mathcal{E}\lossidx{#1}_{\mathrm{E}}}
\newcommand{\dperformance}[1][]{\mathcal{E}\lossidx{#1}_{\mathrm{R}}}
\newcommand{\sg}{\operatorname{sg}}
\newcommand{\ambiguityon}[1]{\ensuremath{\mathtt{on}[#1]}}
\newcommand{\ambiguityoff}[1]{\ensuremath{\mathtt{off}[#1]}} 
\newcounter{contribution}
\newcommand{\contribution}[1]{\refstepcounter{contribution}\label{contribution:#1}\textbf{(C\thecontribution)}}
\newcommand{\contributionref}[1]{\textbf{(C\ref{contribution:#1})}} 

\title{Do Mixture-of-Experts Routers Benefit from Ensembling?}

\author{\name Viktor Vesel\'{y} \email v.vesely@student.rug.nl \\
\addr University of Groningen
\AND
\name Sebastian Jitaru \email s.jitaru@student.rug.nl \\
\addr University of Groningen}

\def\month{MM}
\def\year{YYYY}
\def\openreview{\url{https://openreview.net/forum?id=XXXX}}

\begin{document}

\maketitle

\begin{abstract}
Ensembles of neural networks have repeatedly been shown to outperform single networks. Ensemble theory attributes this gain to the accuracy of the individual members and the diversity among them. Mixture-of-experts models already combine many experts, but a single router decides how the experts are weighted for each input. Motivated by ensemble theory, we apply ensembling to the routing step: an ensemble of routers, trained jointly with the experts, proposes the expert weights. We systematically evaluate this approach using underparameterized networks on two image-classification and two tabular-regression tasks. We analyse each experiment with a performance--diversity decomposition of the router ensemble, which, to our knowledge, has not previously been developed for routers. We report four findings. First, more routers lower the validation loss. Second, training routers to explicitly maximize expert diversity improves the mean performance on three of the four tasks. Third, explicitly maximizing router diversity has no effect on performance. Fourth, combining router weights with the arithmetic mean works better than combining them with the geometric mean. Routing benefits mildly from ensembling, but this benefit is offset by the computational cost of training additional routers.
\end{abstract}

\section{Introduction}

Ensemble methods are a well-established class of machine learning techniques that improve generalization by aggregating the predictions of multiple models. Their effectiveness is typically attributed to two complementary factors: the performance of the individual members and the diversity among them \citep{NIPS1994_b8c37e33, wood2023unified}. Improving individual members lowers their individual error, while diversity promotes error de-correlation, allowing aggregation to cancel out model-specific mistakes.

Classical ensemble methods induce diversity implicitly via bagging \citep{breiman1996bagging}, as in random forests \citep{breiman2001random}, where members are trained on different subsets of the data, whereas boosting methods, such as gradient boosting \citep{friedman2001greedy}, construct members sequentially to focus on errors made by earlier models. In deep learning, diversity can be encouraged explicitly through mechanisms such as negative-correlation learning \citep{chen_ncl_weight}, extended to deep ensembles by \citet{buschjaeger2020generalized}, or expert orthogonalization \citep{liu2024diversifying}. However, \citet{fort2019deep} showed that, in many cases, initializing ensemble members with different random weights is sufficient to induce substantial diversity. Later, it was shown that explicitly maximizing expert diversity can be harmful to overall ensemble performance \citep{wood2023unified}.

Mixture-of-experts models combine several experts with input-dependent weights produced by a learned router (gating network) \citep{jacobs1991adaptive}. Sparse variants, in which only a subset of experts is active for a given input, have gained prominence because of their computational advantages in large-scale language modeling. In the most prominent setting, each expert, or subset of experts, is responsible for a particular region of the input domain, allowing the ensemble to approximate the capacity of a larger model while retaining the computational cost of a smaller one. Existing variants balance computation and model capacity in different ways: selecting the top-$k$ experts for each input \citep{shazeer2017outrageously}, maximizing cost reduction by routing each input to a single expert \citep{fedus2022switch}, or allocating a dynamic number of experts per input \citep{wang2025remoe}. Routing-based methods decompose prediction into two coupled subproblems: a \textit{routing problem}, handled by a \textit{router}, and a \textit{prediction problem}, handled by the \textit{experts}. 

While ensembles typically focus on improving the predictive capacity of the experts, the routing mechanism itself may also benefit from ensembling. This motivates studying whether increasing the number of routers improves the allocation of unseen samples to experts and, consequently, the overall performance of the ensemble. Recent methods combine multiple routers \citep{zhang2025mixture, ran2025router} and tune their number as a hyperparameter. However, to our knowledge, there is no systematic analysis of router ensembles through the lens of ensemble theory. We study this question in a soft mixture-of-experts setting similar to that of \citet{ICLR2024_79fea214}, which combines differentiable routing with soft sparsity.

\paragraph{Contributions} In this work, we provide a systematic study of scaling the number of routers within a mixture-of-experts system using underparameterized MLP architectures. We focus on two image-classification tasks, CIFAR-10 \citep{krizhevsky2009learning} and SVHN \citep{netzer2011reading}, and two tabular-regression tasks, the Bike Sharing dataset \citep{bike_sharing_275} and the Energy Production dataset \citep{energydataset}. Specifically, we identify the following findings:

\begin{itemize}
    \item \contribution{scaling} \textbf{Scaling the number of routers helps mildly}: More routers lower the validation loss on three of the four tasks, by up to about 10\%, with little further improvement beyond roughly 16 routers.
    \item \contribution{predictor-ambiguity} \textbf{Routers benefit from explicitly maximizing expert diversity}: We propose a learning scheme in which routers learn to explicitly maximize expert diversity. Under this scheme, the mean performance is higher on three of the four tasks.
    \item \contribution{delegator-ambiguity} \textbf{Router diversity decomposition}: To our knowledge, we are the first to apply the performance--diversity decomposition to an ensemble of routers. Using this decomposition, we test whether routers also benefit from not explicitly maximizing their diversity. Our findings suggest that, unlike for experts, disabling explicit diversity maximization provides no consistent benefit.
    \item \contribution{aggregation} \textbf{Router aggregation}: We compare arithmetic and normalized geometric means for combining router weights under joint training. The arithmetic mean gives a lower mean validation loss on three of the four tasks.
\end{itemize}

\section{Background} \label{sec:background}

We consider a supervised learning setting with a training dataset $\sD = \{(\vx_n, \vy_n)\}_{n=1}^{N}$ of $N$ samples, which we assume to be drawn i.i.d.\ from an unknown data distribution. Here $\vx \in \sR^{d_x}$ is the input, where $d_x$ is the number of input features. The corresponding output is $\vy \in \sR^{d_y}$ for regression, where $d_y$ is the number of regression targets, or a one-hot target $\vy \in \{0,1\}^{d_y}$ for classification, where $d_y$ is the number of classes.

We define an ensemble of $M$ experts $f_m = f_{\vtheta_m}$, with parameters $\vTheta = (\vtheta_1,\ldots,\vtheta_M)$, and an ensemble of $L$ routers $g_l = g_{\vphi_l}$, with parameters $\vPhi = (\vphi_1,\ldots,\vphi_L)$. We write $\Delta_M = \{\va \in \sR^M : a_m \geq 0,\ \sum_{m=1}^M a_m = 1\}$ for the probability simplex, that is, the set of distributions over the $M$ experts. An expert produces an output vector $\hat{\vy}_m(\vx) = f_m(\vx) \in \sR^{d_y}$. This is a real-valued prediction for regression and a probability vector over the $d_y$ classes for classification. A router produces a weight proposal $\va_l(\vx) = g_l(\vx) \in \Delta_M$, a distribution over the experts that states how much each expert contributes for input $\vx$. Thus, experts estimate the target, while routers determine how experts are combined.

For any weights $\va \in \Delta_M$, define the expert aggregate as prescribed in \citet{wood2023unified}:
\begin{equation}\label{eq:predictor-aggregate}
    F(\vx;\va) =
    \begin{cases}
        \displaystyle\sum_{m=1}^M a_m\hat{\vy}_m(\vx), & \text{regression},\\[2mm]
        \displaystyle\frac{\bigodot_{m=1}^M\hat{\vy}_m(\vx)^{a_m}}
        {\sum_{c=1}^{d_y}\prod_{m=1}^M \hat y_{m,c}(\vx)^{a_m}}, & \text{classification}.
    \end{cases}
\end{equation}
For regression, $F$ is the weighted arithmetic mean of the expert outputs. For classification, $F$ is the weighted geometric mean of the expert probability vectors. The numerator multiplies the expert probabilities, each raised to its weight $a_m$. The denominator rescales the result so that its entries sum to one, which makes $F(\vx;\va)$ a probability vector. In both cases, a larger weight $a_m$ gives expert $m$ more influence on the aggregate, and uniform weights $a_m = 1/M$ give the unweighted mean. Here $\bigodot$ denotes the componentwise product over the experts, and the power $\hat{\vy}_m(\vx)^{a_m}$ is taken componentwise.

The router proposals are combined into $\bar{\va}(\vx) \in \Delta_M$ using either an arithmetic or normalized geometric mean (Section~\ref{sec:exp-agg}). For an input $\vx$, the final prediction is $\hat{\vy} = F(\vx;\bar{\va}(\vx))$. If we aggregate the experts with the weights of a single router $l$ instead, the same function $F$ gives the prediction induced by router $l$ alone, $\hat{\vy}^{(l)} = F(\vx;\va_l(\vx))$. With no routers ($L=0$), we use uniform weights $\bar a_m = 1/M$.

\subsection{Expert performance and diversity}

For a fixed sample $(\vx,\vy)$, write $\hat{\vy}_m=f_m(\vx)$ for each expert's output. Let $\mathrm{SE}(\hat{\vy},\vy)=d_y^{-1}\|\hat{\vy}-\vy\|_2^2$ and $\mathrm{CE}(\hat{\vy},\vy)=-\sum_{c=1}^{d_y}y_c\log\hat y_c$ denote squared error and cross-entropy. For any weights $\va\in\Delta_M$, the classical squared-error decomposition \citep{NIPS1994_b8c37e33} is
\begin{equation}\label{eq:mse-train}
    \mathrm{SE}(F(\vx;\va),\vy)
    = \underbrace{\sum_{m=1}^M a_m\mathrm{SE}(\hat{\vy}_m,\vy)}
        _{\text{weighted expert loss }\pperformance[\mathrm{SE}](\va)}
      -\underbrace{\sum_{m=1}^M a_m\mathrm{SE}(F(\vx;\va),\hat{\vy}_m)}
        _{\text{weighted expert diversity }\pambiguity[\mathrm{SE}](\va)}.
\end{equation}
For cross-entropy with the geometric mean, the corresponding decomposition is \citep{wood2023unified}
\begin{equation}\label{eq:ce-train}
    \mathrm{CE}(F(\vx;\va),\vy)
    = \underbrace{\sum_{m=1}^M a_m\mathrm{CE}(\hat{\vy}_m,\vy)}
        _{\text{weighted expert loss }\pperformance[\mathrm{CE}](\va)}
      -\underbrace{\sum_{m=1}^M a_m\mathrm{KL}(F(\vx;\va)\,\|\,\hat{\vy}_m)}
        _{\text{weighted expert diversity }\pambiguity[\mathrm{CE}](\va)}.
\end{equation}
Here $\mathrm{KL}(\vp\,\|\,\vq)=\sum_{c=1}^{d_y}p_c\log(p_c/q_c)$ is the Kullback--Leibler divergence between two probability vectors $\vp$ and $\vq$.

We use $\ell\in\{\mathrm{SE},\mathrm{CE}\}$ for the task loss and $d^{\ell}$ for the corresponding disagreement measure, with $d^{\mathrm{SE}}=\mathrm{SE}$ and $d^{\mathrm{CE}}=\mathrm{KL}$. Both identities then read
\begin{equation}\label{eq:unified-decomposition}
    \ell(F(\vx;\va),\vy)=\pperformance[\ell](\va)-\pambiguity[\ell](\va),
\end{equation}
with $\pperformance[\ell](\va)=\sum_{m=1}^M a_m\,\ell(\hat{\vy}_m,\vy)$ and $\pambiguity[\ell](\va)=\sum_{m=1}^M a_m\,d^{\ell}(F(\vx;\va),\hat{\vy}_m)$. The loss $\ell$ selects which expressions $\pperformance[\ell]$ and $\pambiguity[\ell]$ stand for, namely those of Equation~\ref{eq:mse-train} or Equation~\ref{eq:ce-train}. When the loss is clear from context, we omit the superscript.

The two terms have a direct meaning. $\pperformance[\ell]$ is the weighted average loss of the individual experts. $\pambiguity[\ell]$ is the weighted average disagreement between the aggregate and the individual experts, and we call it the \emph{diversity} of the experts. \citet{NIPS1994_b8c37e33} named this term the \emph{ambiguity} of the ensemble for squared error, whereas \citet{wood2023unified} call it diversity. We use the term diversity and keep the symbol $\mathcal{A}$ from the original name. Equations~\ref{eq:mse-train}--\ref{eq:ce-train} show how both terms act on the ensemble loss. SE and KL are non-negative, so $\pambiguity[\ell]\geq 0$ and the ensemble loss never exceeds the weighted individual loss $\pperformance[\ell]$. At fixed diversity, every unit by which the individual losses decrease lowers the ensemble loss by one unit. At fixed individual loss, every unit of added diversity lowers the ensemble loss by one unit.

\emph{Reporting convention.} The quantities $\pperformance$ and $\dperformance$ denote losses in all definitions and training equations. In experimental results and plots we talk about \emph{performance}, which is just negative loss $-\pperformance$ and $-\dperformance$, so higher performance is better.

Explicit maximization of diversity has been shown to be potentially harmful, since it can increase the bias and variance of the individual models, leading to higher generalization error \citep{pmlr-v151-ortega22a, wood2023unified}. Prior work also indicates that using different random initializations is often sufficient to induce test-time diversity among ensemble members \citep{fort2019deep, pmlr-v151-ortega22a}. However, it is unknown whether routers benefit from maximizing the expert diversity term. This motivates controlling diversity gradients separately for the routers (Section~\ref{sec:exp-gradient}).

\section{Router ensemble analysis and experiments}

\subsection{Analyzing router ensembles}\label{sec:delegator-decompositions}\label{sec:delegator-ambiguity}

We extend the performance--diversity analysis to the router ensemble. There is no target weight vector for routers. Instead, we assess each weight proposal through the expert ensemble it induces. The loss $\ell(\hat{\vy}^{(l)},\vy)$ measures whether router $l$ proposes weights that yield an accurate prediction. Its diversity $d^{\ell}(\hat{\vy},\hat{\vy}^{(l)})$ measures disagreement with the final prediction, rather than disagreement between weight vectors themselves.

Averaging router weights $\bar{\va}=L^{-1}\sum_{l=1}^L\va_l$ yields a diversity decomposition analogous to Equations~\ref{eq:mse-train}--\ref{eq:ce-train}, but at the router level rather than the expert level:
\begin{equation}\label{eq:delegator-ambiguity}
    \ell(\hat{\vy},\vy)
    = \underbrace{\frac{1}{L}\sum_{l=1}^L\ell(\hat{\vy}^{(l)},\vy)}
        _{\text{individual router loss }\dperformance[\ell]}
      -\underbrace{\frac{1}{L}\sum_{l=1}^L d^{\ell}(\hat{\vy},\hat{\vy}^{(l)})}
        _{\text{router diversity }\dambiguity[\ell]}.
\end{equation}
The derivation for both losses is given in Appendix~\ref{ap:delegator-decomposition}.
Thus, each router should propose weights yielding a good expert ensemble, while diversity is measured in the outputs these proposals induce. A larger $\dambiguity$ is beneficial at fixed $\dperformance$.

Equation~\ref{eq:delegator-ambiguity} does not generally hold for the geometric mean of routing weights. The two terms remain defined, but their difference need not equal the final loss. Both terms require $L\geq 1$.

This decomposition allows us to test whether routers benefit from explicitly maximizing their diversity or whether different random initializations suffice, by disabling gradients through the $\dambiguity$ term.

\subsection{Experimental setup}\label{sec:experimental-setup}

Our claims are validated empirically on four datasets: CIFAR-10 \citep{krizhevsky2009learning}, SVHN \citep{netzer2011reading}, the Bike Sharing dataset \citep{bike_sharing_275}, and the Energy Production dataset \citep{energydataset}. For each dataset, the inputs $\vx$ and targets $\vy$ were normalized where applicable, and categorical integer-valued features were one-hot encoded. For each random seed, the data were independently partitioned into training and validation sets using an $85\%$--$15\%$ split.

These datasets cover a broad range of task difficulty, as shown in Figure~\ref{fig:task-difficulty}. Bikes is the easiest task, followed by SVHN and CIFAR-10, while Energy is the hardest. More importantly, both modalities contain an easier and a harder task: Bikes and Energy for regression, and SVHN and CIFAR-10 for classification. This lets us study the same questions across tasks that differ not only in their data type, but also in how well the tested models can solve them.

For the tabular-regression tasks, Bikes and Energy, we use fully connected MLPs with two hidden layers and optimize them using the mean squared error loss, $\ell_{\mathrm{MSE}}$. The hidden-layer widths decrease progressively toward the output: the two-layer architecture uses widths $(2b, b)$, where $b$ is a model-specific base width. This parameterization preserves a consistent architectural shape while allowing the capacities of the expert and router networks to be varied independently through their respective base widths. The regression models are trained for $2000$ epochs. 

For the image-classification tasks, CIFAR-10 and SVHN, we use three-layer MLPs with hidden widths $(3b, 2b, b)$, where $b$ is a model-specific base width. The classification models are optimized using the cross-entropy loss, $\ell_{\mathrm{CE}}$. Training hyperparameters and experimental parameter ranges are provided in Appendix~\ref{ap:experimental-setup}. The image-classification models are trained for $100$ epochs. 

Following the load-balancing motivation of \citet{shazeer2017outrageously} and \citet{fedus2022switch}, we penalize nonuniform expert usage using Gini impurity. For a minibatch $\sB\subseteq\sD$, the final training objective is
\begin{equation}\label{eq:training-objective}
\begin{aligned} 
    \mathcal{L}_{\sB}
    &= \frac{1}{|\sB|}\sum_{(\vx,\vy)\in\sB}\ell(\vx,\vy)
       +\lambda\frac{\sum_{m=1}^M u_m^2-1/M}{1-1/M},\\
    \vu &= \frac{1}{|\sB|}\sum_{(\vx,\vy)\in\sB}\bar{\va}(\vx),
\end{aligned}
\end{equation} 
where $M>1$, $\lambda=0.2$, and $\ell_{\mathrm{train}}$ has the value of the ensemble loss but the selected training gradients defined in Section~\ref{sec:exp-gradient}. This penalty encourages balanced expert use across the minibatch.

\paragraph{How we analyze the experiments.} We organize the analysis around three questions:
\begin{enumerate}
    \item \textbf{Did the full system get better?} We compare the validation performance of the complete expert and router ensemble. Performance denotes negative task loss $-\ell$, so higher is better.
    \item \textbf{What happened to the individual ensemble members?} The expert performance--diversity decomposition ($-\pperformance$ and $\pambiguity$) describes how well individual experts perform and how different their predictions are. The router decomposition ($-\dperformance$ and $\dambiguity$) describes the same properties for the predictions induced by individual routers' weights.
    \item \textbf{How many experts are utilized?} Local usage measures mean effective expert use per validation sample; global usage measures it across the validation set. Both are entropy-based percentages. Their precise definitions are in Appendix~\ref{ap:usage-definition}.
\end{enumerate}
Each experiment's main-text discussion focuses on the analyses most relevant to its question; the remaining decomposition and usage results are collected in Appendix~\ref{ap:additional-results}, organized in the same order as the experiments below.

Each experimental run is evaluated across five independent random seeds, which control network initialization, the data split, and minibatch order. We evaluate the checkpoint selected by early stopping and average the resulting metrics over seeds for each run. We conduct two types of experiments: scaling (Section~\ref{sec:exp-scaling}) and paired comparisons (Sections~\ref{sec:exp-gradient}--\ref{sec:exp-agg}). In the scaling experiments, we vary the number of routers $L$ while holding all other variables fixed, and exhaustively evaluate the remaining experimental configurations. In the paired experiments, we randomly sample configurations of the remaining variables, including the seed, and evaluate both values of the variable of interest under identical configurations.

\subsection{Scaling the number of routers} \label{sec:exp-scaling}

In this experiment, we investigate whether routing itself benefits from an ensemble effect (Contribution~\contributionref{scaling}). To isolate the effect of the number of routers, we vary the number of routers $L \in \{0,1,2,4,8,16,32\}$ while evaluating all combinations of the number of experts $M \in \{2,4,8,16,32\}$, expert base width $b_M \in \{4,8,16\}$, and router base width $b_L \in \{4,8,16\}$ across all four tasks. The case $L=0$ corresponds to fixed uniform routing weights $\bar{\va}$, whereas $L=1$ recovers a setting closely related to a standard mixture-of-experts model. For each combination of the other variables, we use the same five random seeds across all values of $L$, but a different set of five seeds for each combination.

\paragraph{Ensemble performance}  

\begin{figure}[h]
    \centering
    \includegraphics[width=0.99\linewidth]{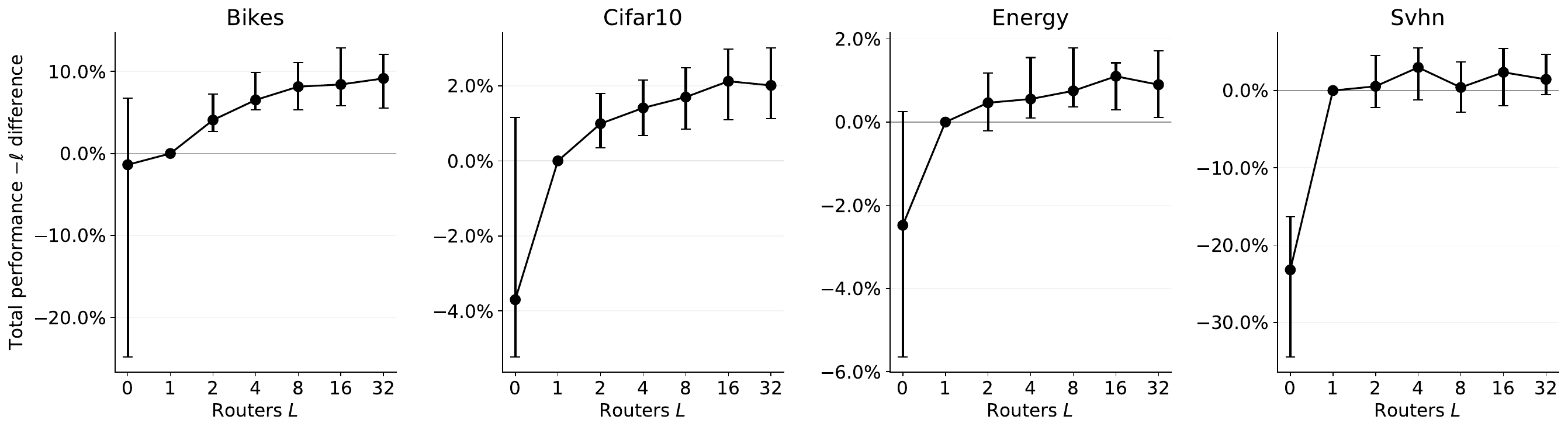}
    \caption{Number of routers scaling experiment: The total ensemble benefits mildly from the addition of routers on three of the four tasks. Relative validation-performance difference for $L \in \{0,1,2,4,8,16,32\}$ compared with the single-router setting, $L=1$, computed on validation samples. The $L=0$ setting uses fixed uniform weights, while $L=1$ is closely related to a standard mixture-of-experts model. The curves show the mean, with error bars indicating the interquartile range.}
    \label{fig:scaling-validation} 
\end{figure}

The total ensemble benefits mildly from the addition of routers. Figure~\ref{fig:scaling-validation} shows a consistent but modest benefit from adding routers on Bikes, CIFAR-10, and Energy, with little further improvement beyond roughly $L=16$. Bikes shows the clearest gain, around $10\%$ of the baseline loss magnitude; gains on CIFAR-10 and Energy are smaller. SVHN shows a clear benefit from having at least one router (a $20\%$ improvement from $L=0$ to $L=1$), but no consistent gains from adding further routers. Across all four tasks, fixed uniform weights without a router ($L=0$) yield worse ensemble performance than learned routing.

\paragraph{Experts' performance} 

\begin{figure}[h]
    \centering
    \includegraphics[width=0.99\linewidth]{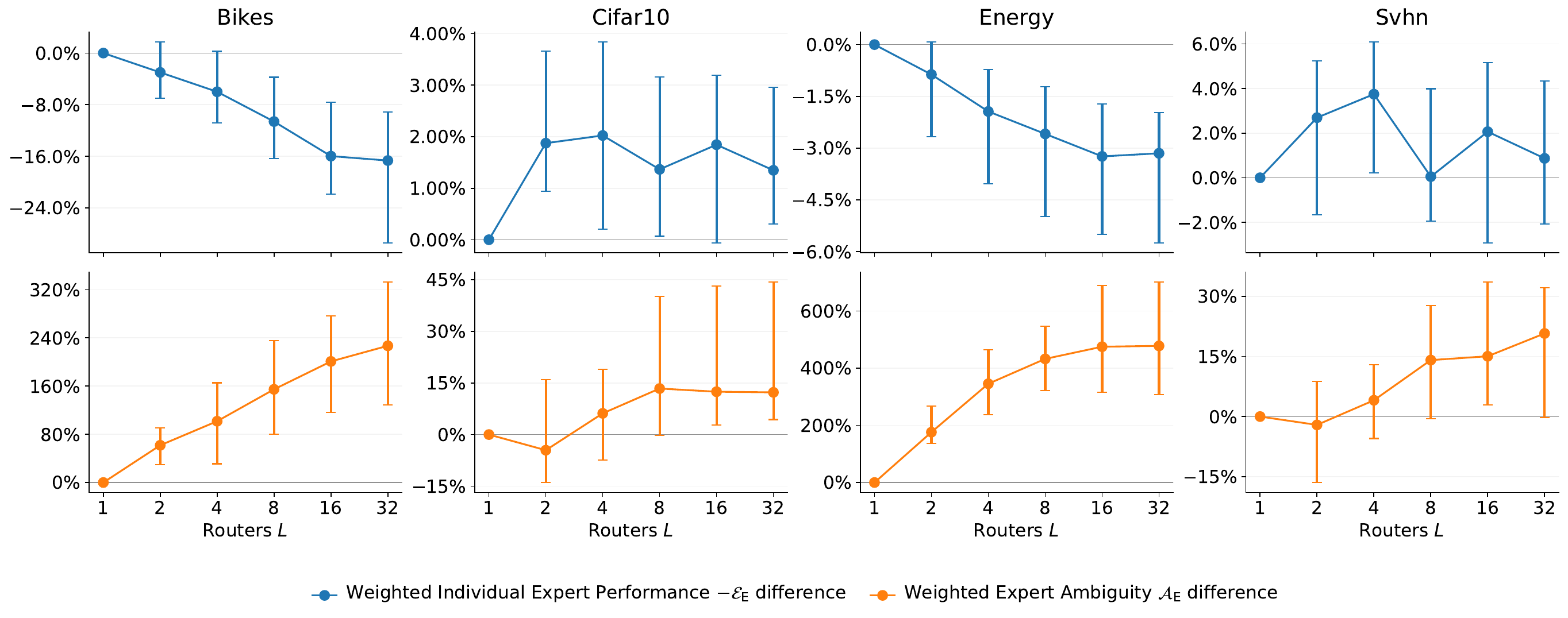}
    \caption{Number of routers scaling experiment: Adding more routers consistently increases diversity among the experts, but this positive effect is partially offset by a deterioration in individual expert performance. Relative differences in expert performance (top row) and expert diversity (bottom row) for $L \in \{1,2,4,8,16,32\}$ compared with $L=1$, computed on validation samples. Each column shows one task. The curves show the mean, with error bars indicating the interquartile range.}
    \label{fig:scaling-predictors}
\end{figure}

Adding more routers consistently increases diversity among the experts, but this benefit is partially offset by a deterioration in individual expert performance. Figure~\ref{fig:scaling-predictors} shows increasing expert diversity, particularly on regression tasks, accompanied by a general increase in individual expert error. This deterioration cannot be readily attributed to explicitly encouraging routers to identify diverse experts through maximization of the expert diversity term. As shown in Section~\ref{sec:exp-gradient}, enabling this term instead improves both overall ensemble performance and individual expert performance. The increase in individual expert error therefore appears to be associated with scaling the number of routers itself, rather than explicitly maximizing expert diversity. Understanding and mitigating this effect could help translate the increased expert diversity into larger improvements in overall ensemble performance.

\paragraph{Routers' performance}

\begin{figure}[h]
    \centering
    \includegraphics[width=0.99\linewidth]{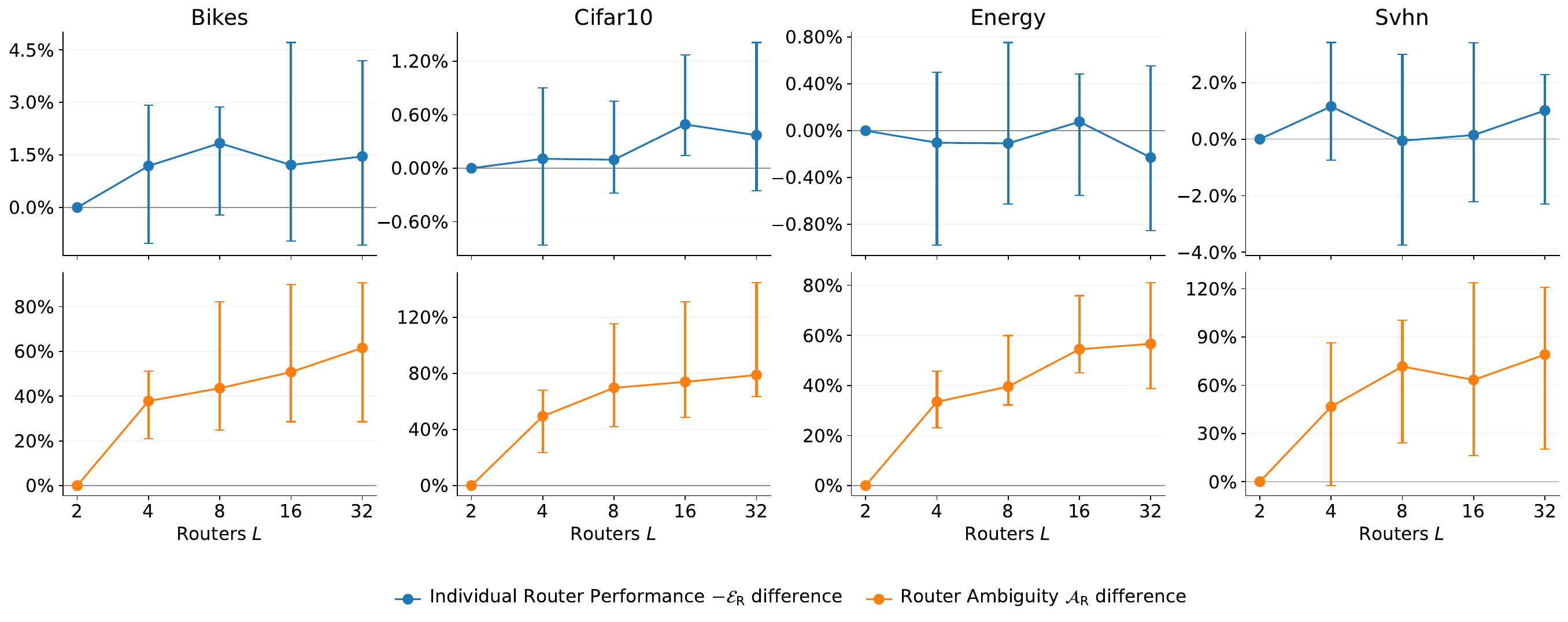}
    \caption{Number of routers scaling experiment: Scaling the number of routers consistently improves the diversity of the ensemble they induce, while individual router performance remains constant. Relative differences in router performance (top row) and router diversity (bottom row) for $L \in \{2,4,8,16,32\}$ compared with $L=2$, computed on validation samples. The $L=0$ setting is omitted because it has no learned routers, and $L=1$ is omitted because a single router has zero router diversity. Each column shows one task. The curves show the mean, with error bars indicating the interquartile range.}
    \label{fig:scaling-delegators}
\end{figure}

Scaling the number of routers consistently improves the diversity of the ensemble they induce, while individual router performance remains constant (Figure~\ref{fig:scaling-delegators}). This means that each router individually proposes accurate weights even in large router ensembles. Thus, the decrease in individual expert performance is probably caused by the routing of samples to experts during training, leading to some form of overfitting, rather than by inaccurate routers.

\paragraph{Expert usage}

\begin{figure}[h]
    \centering
    \includegraphics[width=0.99\linewidth]{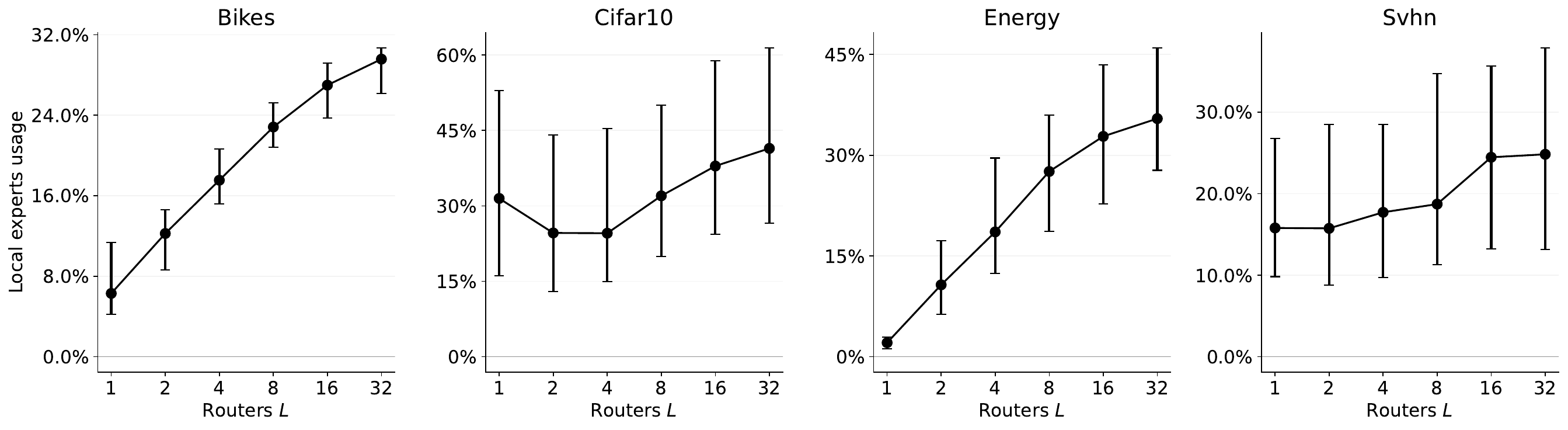}
    \caption{Number of routers scaling experiment: Local expert usage increases with the number of routers. Absolute local expert usage for $L \in \{1,2,4,8,16,32\}$, computed on validation samples and reported as a percentage. Each column shows one task. The curves show the mean, with error bars indicating the interquartile range.}
    \label{fig:scaling-local-usage}
\end{figure}

Local expert usage increases with the number of routers, providing another perspective on the total ensemble improvement. Figure~\ref{fig:scaling-local-usage} shows that the increase in usage is especially strong on Bikes and Energy. This makes intuitive sense, as averaging more router proposals should lead to more uniform weights. However, this result does not suggest that more uniform usage leads to a better ensemble: in the no-router setting ($L=0$), usage is $100\%$ (fully uniform), yet the total ensemble performance is consistently lower.

%
%
%

\subsection{Expert diversity gradient} \label{sec:exp-gradient}

As discussed in Section~\ref{sec:background}, explicitly maximizing expert diversity can harm individual expert performance. We therefore stop the expert diversity gradient from influencing the expert parameters $\vTheta$ in both conditions. The experiment changes only whether the routers learn from the expert diversity term $\pambiguity$:
\begin{equation}\label{eq:pred-ambiguity-choices}
    \underbrace{\pperformance(\bar{\va})-\pambiguity(\bar{\va})}
        _{\ambiguityon{\pambiguity}}
    \qquad\text{or}\qquad
    \underbrace{\pperformance(\bar{\va})}
        _{\ambiguityoff{\pambiguity}}.
\end{equation}
In the \ambiguityoff{\pambiguity} condition, routers only learn to identify correct experts, while in the \ambiguityon{\pambiguity} condition, they can also minimize the loss by identifying decorrelated experts. We use a paired design by sampling 100 base configurations and comparing \ambiguityon{\pambiguity} with \ambiguityoff{\pambiguity} in each (Contribution~\contributionref{predictor-ambiguity}). The full learning equations, formulated using the stop-gradient operator, are given in Appendix~\ref{ap:gradient-routing}.

\paragraph{Ensemble performance} 

\begin{figure}[h]
    \centering
    \includegraphics[width=0.99\linewidth]{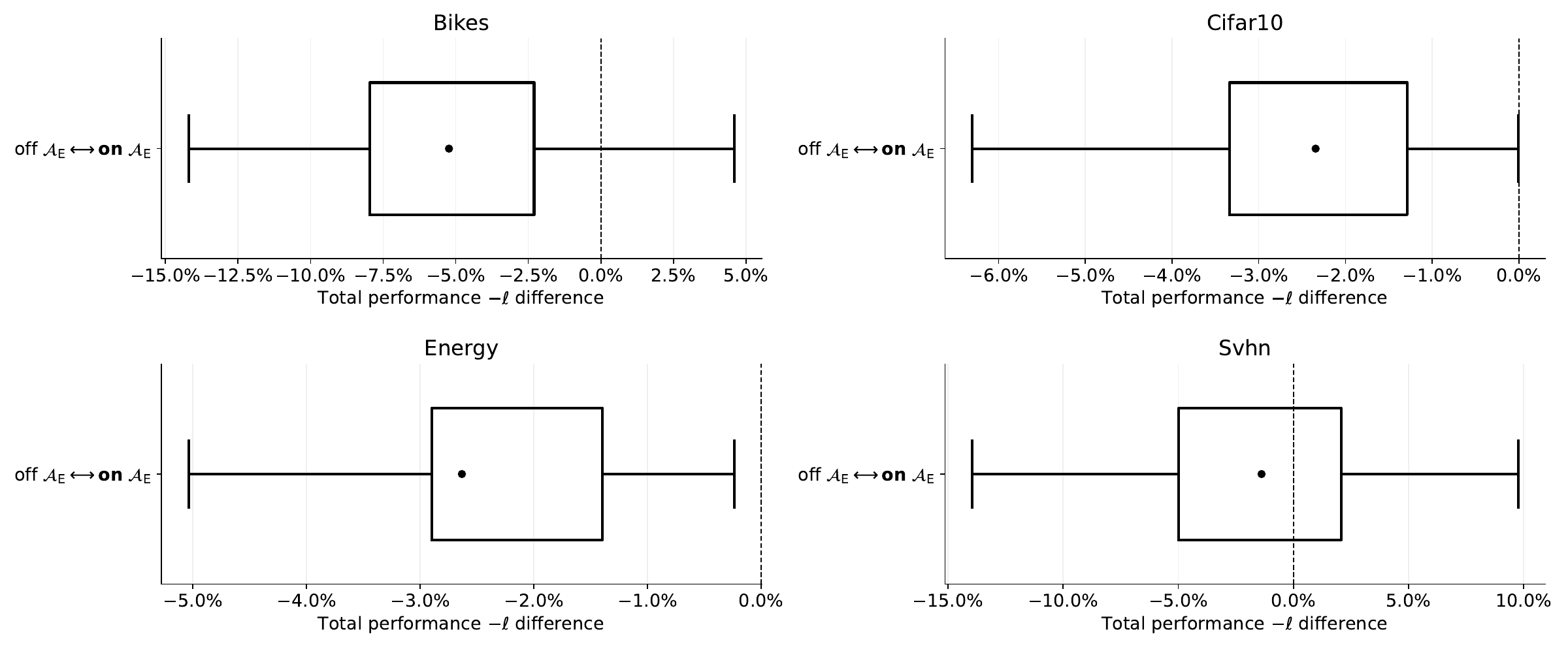}
    \caption{Expert diversity gradient experiment: Distribution of the relative validation-performance differences between \ambiguityon{\pambiguity} and \ambiguityoff{\pambiguity}, computed on validation samples. The better condition is marked in bold on the axis. The black dot denotes the mean relative difference.}
    \label{fig:predictor-gradient-validation}
\end{figure}

Enabling the routers to learn to explicitly identify decorrelated experts increases total ensemble performance. Figure~\ref{fig:predictor-gradient-validation} shows an average performance increase of $5\%$ with \ambiguityon{\pambiguity} versus \ambiguityoff{\pambiguity} on all tasks except SVHN. The reasons are identified in the next paragraph.

\paragraph{Experts' performance}

\begin{figure}[h]
    \centering
    \includegraphics[width=0.99\linewidth]{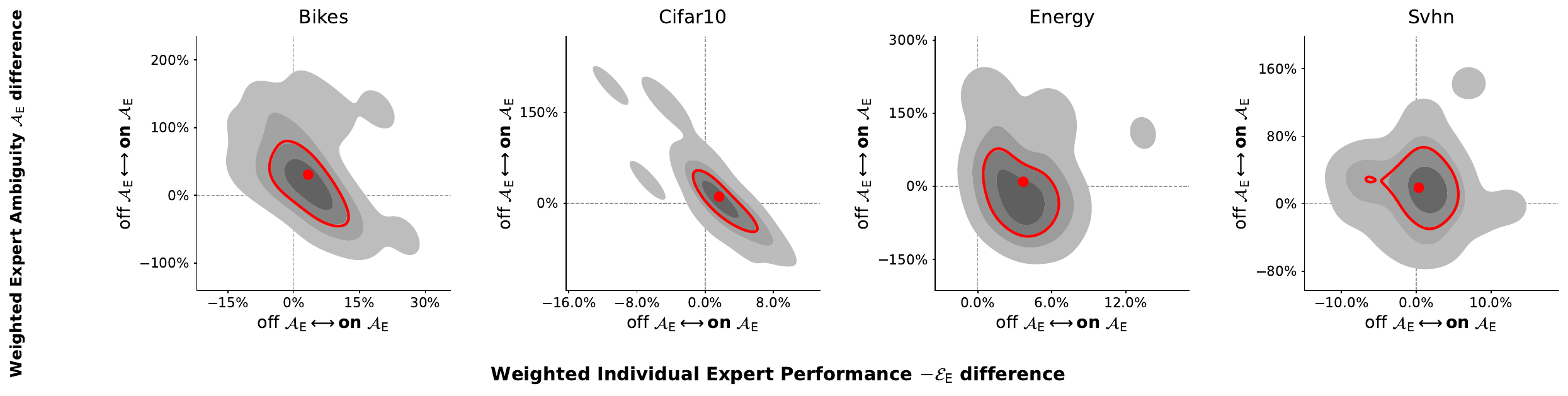}
    \caption{Expert diversity gradient experiment: Experts are more accurate but only consistently more diverse in the Bikes and SVHN tasks. Distribution of the expert performance differences (x-axis) and expert diversity differences (y-axis) between \ambiguityon{\pambiguity} and \ambiguityoff{\pambiguity}, computed on validation samples. The better condition is marked in bold separately on each axis. The red dot denotes the two-dimensional mean, and the red contour encloses the central $50\%$ of the joint distribution.}
    \label{fig:predictor-gradient-predictors}
\end{figure}

Interestingly, the \ambiguityon{\pambiguity} condition produced experts that are more individually accurate and only sometimes more diverse. As seen in Figure~\ref{fig:predictor-gradient-predictors}, the \ambiguityon{\pambiguity} condition produced more diverse experts only on the SVHN and Bikes tasks. The main benefit seems to come from identifying more accurate experts. Surprisingly, however, the diversity benefit of \ambiguityon{\pambiguity} is reflected in router diversity, with individual routers proposing decorrelated ensemble decisions. The mechanism linking \ambiguityon{\pambiguity} to increased router diversity is unclear, but it may be attributed to the additional learning signal from the $\pambiguity$ term, which leads the individual router parameters $\vphi_i$ to converge to different solutions.

\subsection{Router diversity gradient} \label{sec:exp-delegator-gradient}

We next test whether routers should learn to explicitly maximize their diversity $\dambiguity$. We compare two conditions. In \ambiguityoff{\dambiguity}, routers only learn to propose weights leading to accurate ensemble decisions, whereas in \ambiguityon{\dambiguity}, they also learn to propose weights leading to decorrelated ensemble decisions:
\begin{equation}\label{eq:del-ambiguity-choices}
    \underbrace{\dperformance(\bar{\va})-\dambiguity(\bar{\va})}
        _{\ambiguityon{\dambiguity}}
    \qquad\text{or}\qquad
    \underbrace{\dperformance} 
        _{\ambiguityoff{\dambiguity}}.
\end{equation}
All other task-loss gradients to routers are stopped; the load-balancing penalty remains active in both conditions.

To isolate the effect of router diversity for Contribution~\contributionref{delegator-ambiguity}, we use a paired design by sampling 100 base configurations and comparing \ambiguityon{\dambiguity} with \ambiguityoff{\dambiguity} in each. We keep expert diversity fixed at \ambiguityon{\pambiguity}.

\paragraph{Ensemble performance}

\begin{figure}[h]
    \centering
    \includegraphics[width=0.99\linewidth]{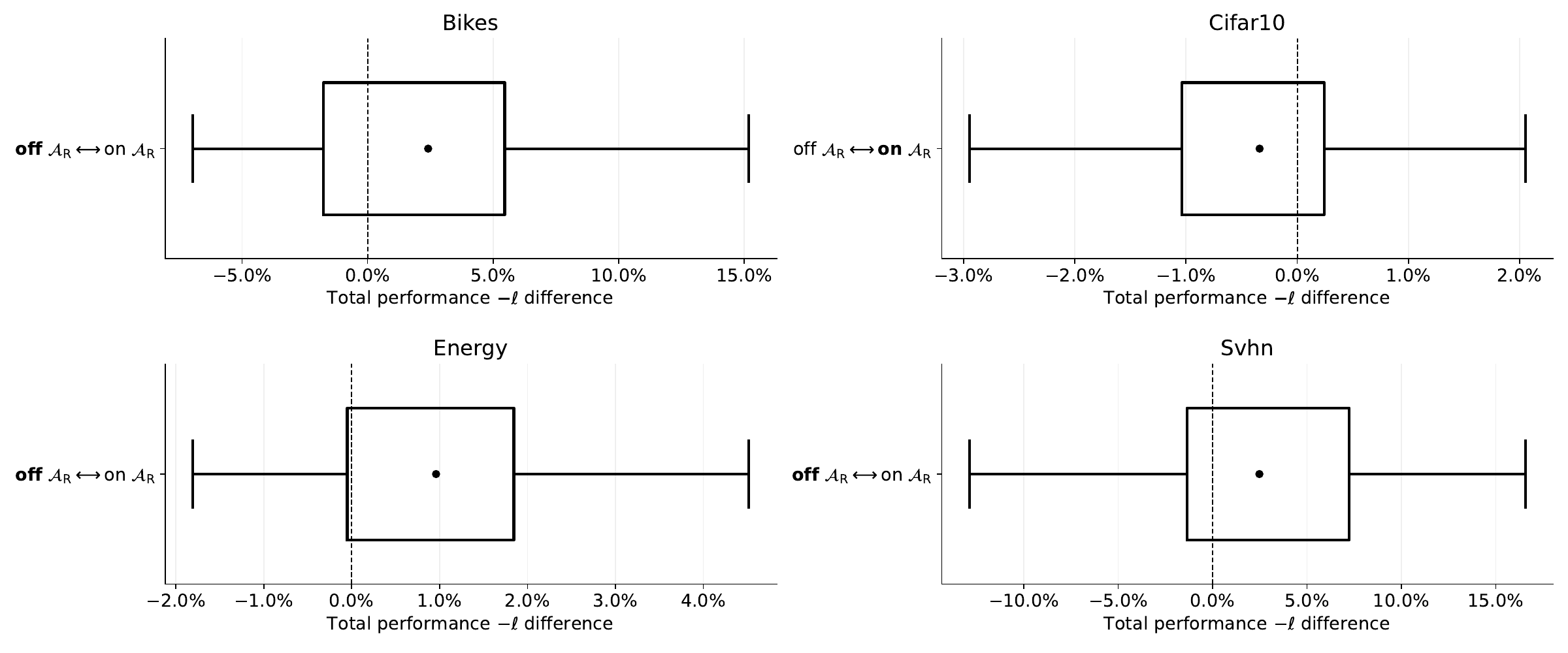}
    \caption{Router diversity gradient experiment: Distribution of the relative validation-performance differences between \ambiguityon{\dambiguity} and \ambiguityoff{\dambiguity}, computed on validation samples. The better condition is marked in bold on the axis. The black dot denotes the mean relative difference.}
    \label{fig:delegator-gradient-validation}
\end{figure}

The \ambiguityoff{\dambiguity} condition leads to slightly better ensembles, but the improvement is inconsistent. Figure~\ref{fig:delegator-gradient-validation} shows a small mean improvement only on the Bikes task. For the other tasks, the improvement is not significant, suggesting that, unlike the findings for deep expert ensembles \citep{fort2019deep}, explicitly maximizing router diversity does not seem to harm performance, at least for the number of parameters and network architectures considered here.

\paragraph{Routers' performance}

\begin{figure}[h]
    \centering
    \includegraphics[width=0.99\linewidth]{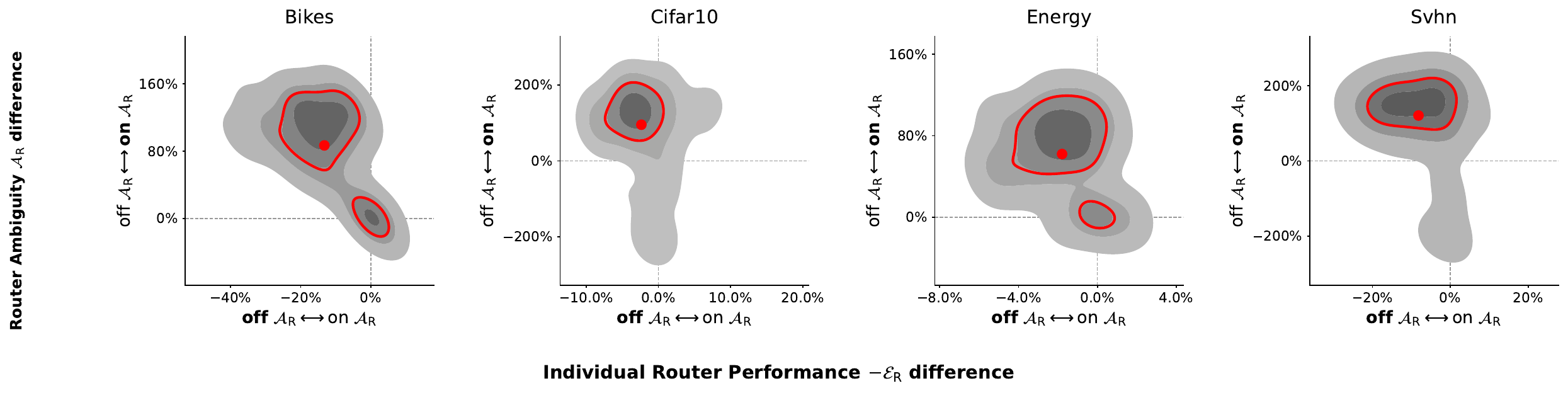}
    \caption{Router diversity gradient experiment: Explicitly maximizing router diversity reduces individual router accuracy but increases diversity; the effects mostly cancel out. Distribution of the router performance differences (x-axis) and router diversity differences (y-axis) between \ambiguityon{\dambiguity} and \ambiguityoff{\dambiguity}, computed on validation samples. The better condition is marked in bold separately on each axis. The red dot denotes the two-dimensional mean, and the red contour encloses the central $50\%$ of the joint distribution.}
    \label{fig:delegator-gradient-delegators}
\end{figure}

Explicitly maximizing router diversity reduces individual router accuracy, but the resulting increase in diversity compensates for this deterioration, unlike in the case of experts. Figure~\ref{fig:delegator-gradient-delegators} shows that these findings are consistent across all tasks. Interestingly, for the regression tasks (Bikes, Energy), a considerable number of runs show no effect from switching between conditions, indicating two different modes of ensemble behaviour.

\subsection{Router aggregation method} \label{sec:exp-agg}

Previous work \citep{pmlr-v151-ortega22a, wood2023unified} characterizes aggregation rules for experts under a fixed set of weights $\va$. In our setting, expert and router parameters are optimized jointly, so the aggregation rule also affects how the two ensembles co-adapt during training. We therefore compare the two aggregation rules empirically, corresponding to Contribution~\contributionref{aggregation}.

We compare two aggregation rules: the sum  $\bar{\va}(\vx) = L^{-1}\sum_{l=1}^{L}\va_l(\vx)$ (arithmetic mean), and the product, $\bar{\va}(\vx) = Z^{-1}\bigodot_{l=1}^{L}\va_l(\vx)^{L^{-1}}$ (normalized geometric mean), where $Z$ normalizes the weights to sum to one. The expert aggregation in Equation~\ref{eq:predictor-aggregate} remains unchanged. We use the \ambiguityoff{\pambiguity} setting (Section~\ref{sec:exp-gradient}) and the \ambiguityoff{\dambiguity} setting (Section~\ref{sec:exp-delegator-gradient}).

To isolate the effect of the aggregation rule, we use a paired design by sampling 100 base configurations and comparing the arithmetic and geometric means in each. 
 
\paragraph{Ensemble performance}

\begin{figure}[h]
    \centering
    \includegraphics[width=0.99\linewidth]{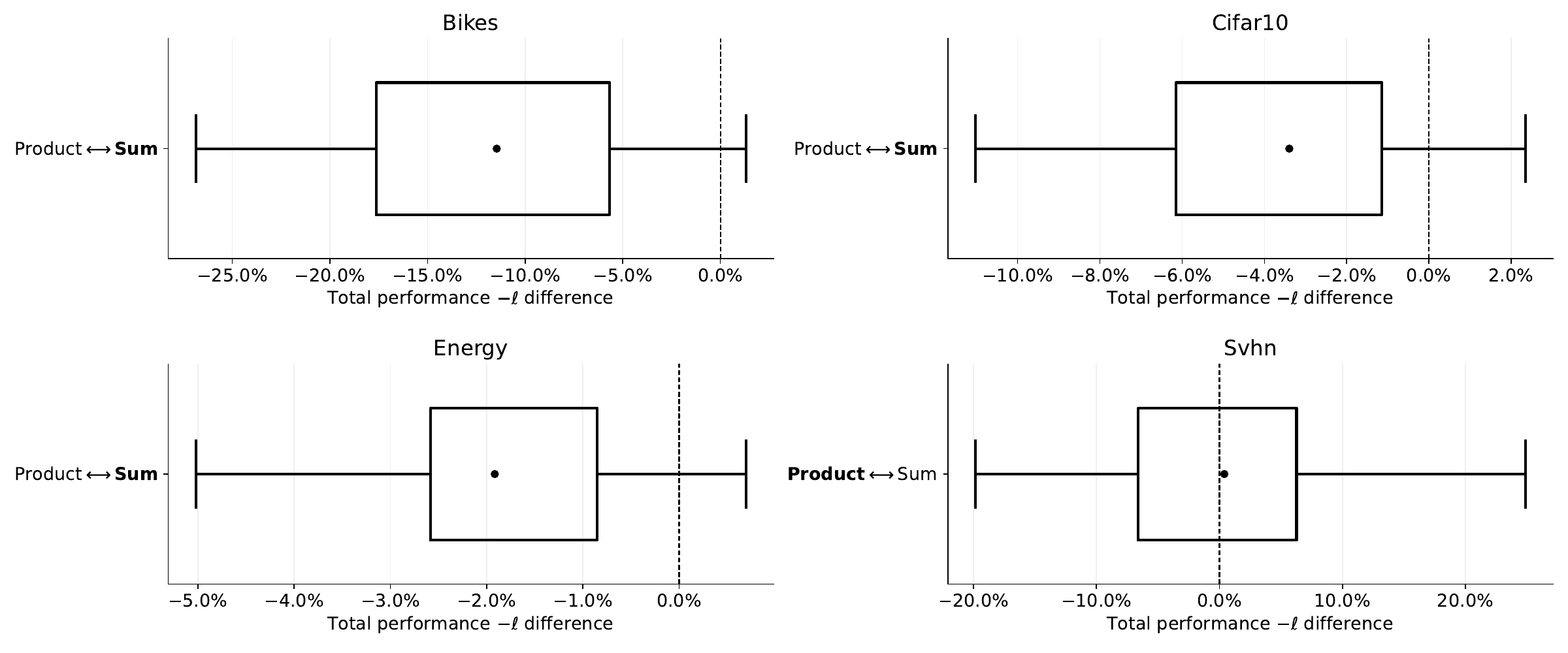}
    \caption{Router aggregation experiment: The sum rule outperforms the product rule on three of the four tasks. Distribution of the relative validation-performance differences between arithmetic and normalized geometric means, computed on validation samples. The better aggregation rule is marked in bold on the axis. The black dot denotes the mean relative difference.}
    \label{fig:aggregation-validation}
\end{figure}

The sum rule outperforms the product rule on three of the four tasks, although the magnitude of the improvement varies. As shown in Figure~\ref{fig:aggregation-validation}, Bikes exhibits the largest improvement, averaging $12\%$, while CIFAR-10 and Energy show more modest gains of approximately $2\%$. SVHN presents a less consistent pattern, with the sum rule substantially improving performance in some runs while degrading it in others, resulting in an interquartile range spanning approximately $-20\%$ to $20\%$. One possible explanation for the worse performance of the product rule is that it underutilizes experts for individual validation samples, as shown by the local usage in Figure~\ref{fig:delegator-gradient-local-usage}. However, the global usage in Figure~\ref{fig:delegator-gradient-global-usage} indicates that all experts are utilized approximately uniformly across the entire validation set.

%
%


\section{Conclusion}\label{sec:discussion}

The weaker performance of the no-router setting ($L=0$) in these softly sparse ensembles motivates learned routing for predictive performance, beyond its computational role in sparse expert selection. Ensemble theory predicts a benefit from greater diversity when individual performance is maintained \citep{NIPS1994_b8c37e33, wood2023unified}. Scaling the number of routers approximately follows this pattern: the predictions induced by individual routers become more diverse while their mean performance changes little. The resulting ensemble gains are modest, occur on three of the four tasks, and largely plateau beyond $L=16$. Further improvements may come from identifying and mitigating the causes of declining individual expert performance as the number of routers increases, because expert diversity, router performance, and router diversity remain strong.

Prior work cautions that explicitly maximizing expert diversity can harm individual expert performance \citep{pmlr-v151-ortega22a, wood2023unified}. Here, we find that allowing only the routers to maximize expert diversity improves individual expert performance but has an inconsistent effect on expert diversity. Instead, the extra learning signal provides a consistent diversity benefit to the routers themselves.

Directly maximizing router decorrelation produces an accuracy--diversity tradeoff: individual routers become less accurate, but the increase in their diversity largely compensates for this deterioration; thus, their total ensemble performance is not affected. Previous work also suggests that different random initializations can provide sufficient expert diversity without explicitly rewarding it \citep{fort2019deep}; our results do not show the same effect for routers.

The arithmetic mean outperforms the geometric mean on three of the four tasks, possibly because it distributes weight across more experts per sample. The geometric mean concentrates weight on very few experts, approaching single-expert selection on regression tasks, while maintaining broad expert usage across the dataset. Its weaker performance may therefore reflect limited use of complementary experts within individual predictions rather than neglecting particular experts altogether. However, fixed uniform routing maximizes local usage yet underperforms learned routing, suggesting that the benefit depends on learning which experts to combine rather than simply using more experts.

For the tasks and architectures presented here, the improvements are modest and probably do not justify training and evaluating many additional models, particularly beyond the plateau, although runtime was not measured here.

\subsubsection*{Acknowledgments}
We thank Professor Davide Grossi for his supervision, guidance, and valuable feedback throughout this project.

\clearpage
\bibliography{tmlr-template/tmlr}
\bibliographystyle{tmlr-template/tmlr}

\clearpage
\appendix

\raggedbottom
\section*{Appendix}

\section{Experimental setup}\label{ap:experimental-setup}

\subsection{Training specifications} \label{ap:hyperparams}

\begingroup
\setlength{\intextsep}{6pt}
\begin{table}[H]
\caption{Training and initialization hyperparameters used across experiments.}
\label{tab:training-hyperparameters}
\centering
\begin{tabular}{ll}
\multicolumn{1}{c}{\bf HYPERPARAMETER}
&
\multicolumn{1}{c}{\bf VALUE}
\\ \hline \\
Kernel initialization          & LeCun normal \\
Bias initialization            & Zero \\
Hidden-layer activation        & ReLU \\
Optimizer                      & Adam \\
Learning rate                  & $10^{-3}$ \\
Weight-decay coefficient       & $10^{-4}$ \\ 
Tabular-task batch size        & $256$ \\
Image-task batch size          & $64$ \\
Tabular-task training epochs   & $2000$ \\
Image-task training epochs     & $100$ \\
\end{tabular}
\end{table} 

\begin{table}[H]
\caption{Experimental parameter ranges.}
\label{tab:experiment-parameters}
\centering
\begin{tabular}{lll}
\multicolumn{1}{c}{\bf VARIABLE}
&
\multicolumn{1}{c}{\bf SET}
&
\multicolumn{1}{c}{\bf NOTES}
\\ \hline \\
Number of experts ($M$) & \{$2, 4, 8, 16, 32$\} &  \\
Number of routers ($L$) & \{$0, 1, 2, 4, 8, 16, 32$\} &  $L=0$: uniform weights; $L=1$: MoE. \\
Experts' base width ($b_M$) & \{$4, 8, 16$\} & Widths: $(2b,b)$ or $(3b,2b,b)$.  \\
Routers' base width ($b_L$) & \{$4, 8, 16$\} & Widths: $(2b,b)$ or $(3b,2b,b)$. \\
Routers' aggregation method & \{arithmetic mean, geometric mean\} & See Section \ref{sec:exp-agg} \\
Expert diversity gradient to routers & \ambiguityoff{\pambiguity}, \ambiguityon{\pambiguity} & Section~\ref{sec:exp-gradient}. \\
Router diversity gradient to routers & \ambiguityoff{\dambiguity}, \ambiguityon{\dambiguity} & Section~\ref{sec:exp-delegator-gradient}. \\
\end{tabular}
\end{table}
\endgroup

\subsection{Expert usage}\label{ap:usage-definition}

For a routing-weight vector $\va\in\Delta_M$ with $M>1$, define normalized entropy and normalized effective usage by
\begin{equation}
    H(\va)=-\frac{\sum_{m=1}^M a_m\log a_m}{\log M},
    \qquad U(\va)=\frac{M^{H(\va)}-1}{M-1},
\end{equation}
with $0\log 0=0$. On a validation set of size $N_{\mathrm{val}}$, local and global usage are
\begin{equation}
    U_{\mathrm{local}}=\frac{1}{N_{\mathrm{val}}}\sum_{n=1}^{N_{\mathrm{val}}}U(\bar{\va}(\vx_n)),
    \qquad
    U_{\mathrm{global}}=U\!\left(\frac{1}{N_{\mathrm{val}}}\sum_{n=1}^{N_{\mathrm{val}}}\bar{\va}(\vx_n)\right).
\end{equation}
We report both as percentages. Zero local usage means that each sample assigns all weight to one expert; zero global usage means that the same expert receives all weight across the validation set. Full local usage means uniform weights for every sample; full global usage means uniform average weights across the validation set. These measures quantify weight concentration, not the fraction of experts executed. We compute them separately for each seed and then average over seeds.

\subsection{Task difficulty}

\begin{figure}[H]
    \centering
    \includegraphics[width=0.8\linewidth]{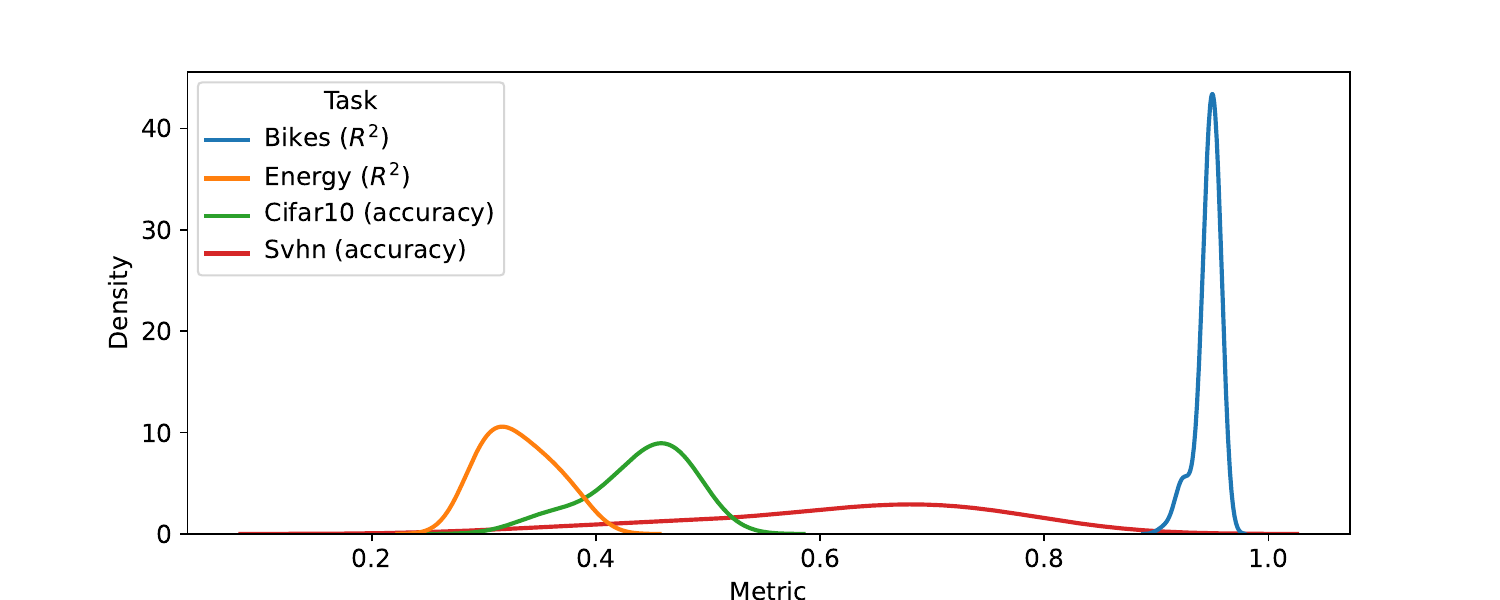}
    \caption{Distribution of absolute validation performance across the four tasks. Performance is measured using accuracy for the image-classification tasks, CIFAR-10 and SVHN, and $R^2$ for the regression tasks, Bikes and Energy. Both measures share the same axis because they have a similar scale in these experiments. The four tasks cover a broad range from easier to harder problems within and across the two modalities.}
    \label{fig:task-difficulty}
\end{figure}

\section{Additional experimental results}\label{ap:additional-results}

\subsection{Router scaling}\label{ap:scaling-global-usage}

\paragraph{Expert usage.}

\begin{figure}[H]
    \centering
    \includegraphics[width=0.99\linewidth]{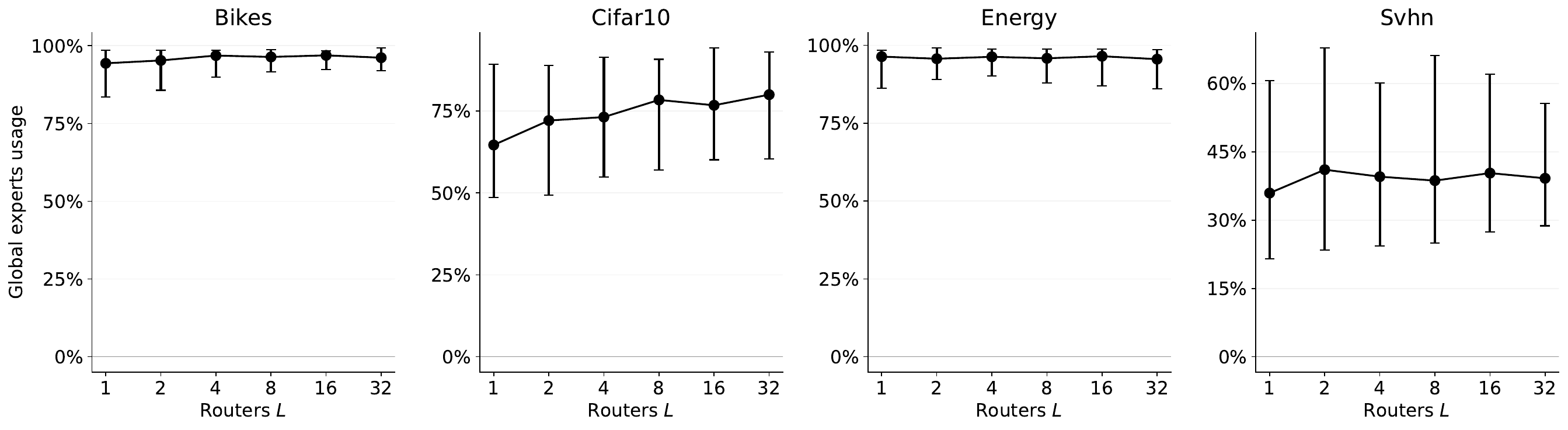}
    \caption{Number of routers scaling experiment: Absolute global expert usage for $L \in \{1,2,4,8,16,32\}$, computed on validation samples and reported as a percentage. Each column shows one task. The curves show the mean, with error bars indicating the interquartile range.}
    \label{fig:scaling-global-usage}
\end{figure}

Global usage remains near its maximum on the regression tasks, whereas it is lower on CIFAR-10 and especially SVHN (Figure~\ref{fig:scaling-global-usage}). Low local usage can thus coexist with broad usage across samples. Lower validation usage on image tasks might reflect patterns that do not transfer from training, or a load-balancing penalty too weak relative to cross-entropy; distinguishing these explanations requires further evidence.

\subsection{Expert diversity}\label{ap:predictor-gradient-delegators}

\paragraph{Router decomposition.}

\begin{figure}[H]
    \centering
    \includegraphics[width=0.99\linewidth]{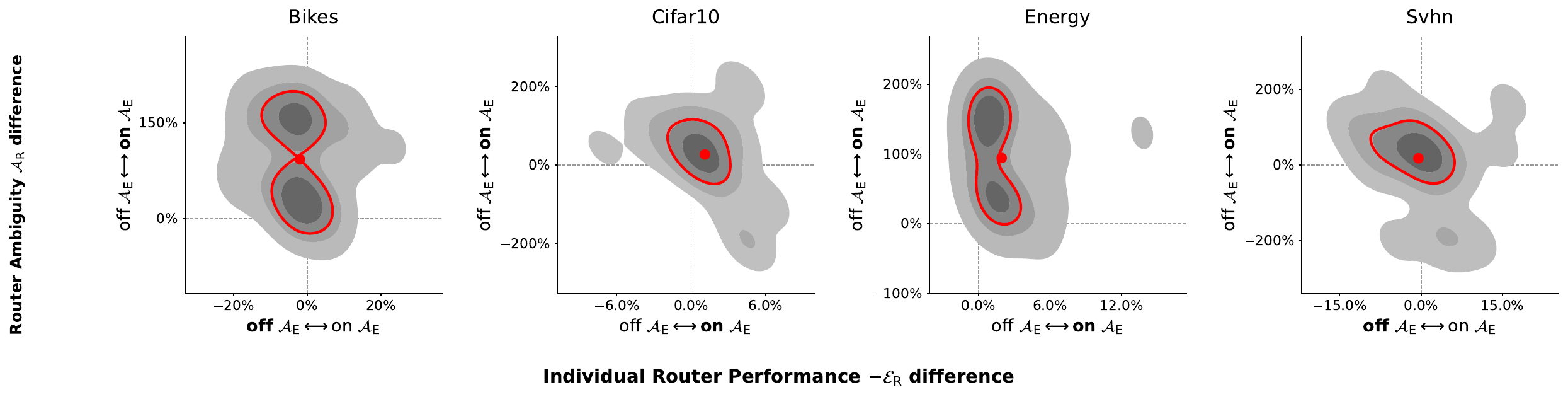}
    \caption{Expert diversity gradient experiment: Distribution of the router performance differences (x-axis) and router diversity differences (y-axis) between \ambiguityon{\pambiguity} and \ambiguityoff{\pambiguity}, computed on validation samples. The better condition is marked in bold separately on each axis. The red dot denotes the two-dimensional mean, and the red contour encloses the central $50\%$ of the joint distribution.} 
    \label{fig:predictor-gradient-delegators}
\end{figure}

The clearer accompanying change is greater router diversity on every task (Figure~\ref{fig:predictor-gradient-delegators}). Individual router performance is approximately balanced on Bikes, CIFAR-10, and SVHN, while Energy also shows an improvement. Thus, a learning signal defined through expert complementarity changes the diversity of router-induced predictions; joint training prevents attributing the gain solely to a change in routing over fixed experts.

\paragraph{Expert usage.}\label{ap:usage-predictor}

\begin{figure}[H]
    \centering
    \includegraphics[width=0.99\linewidth]{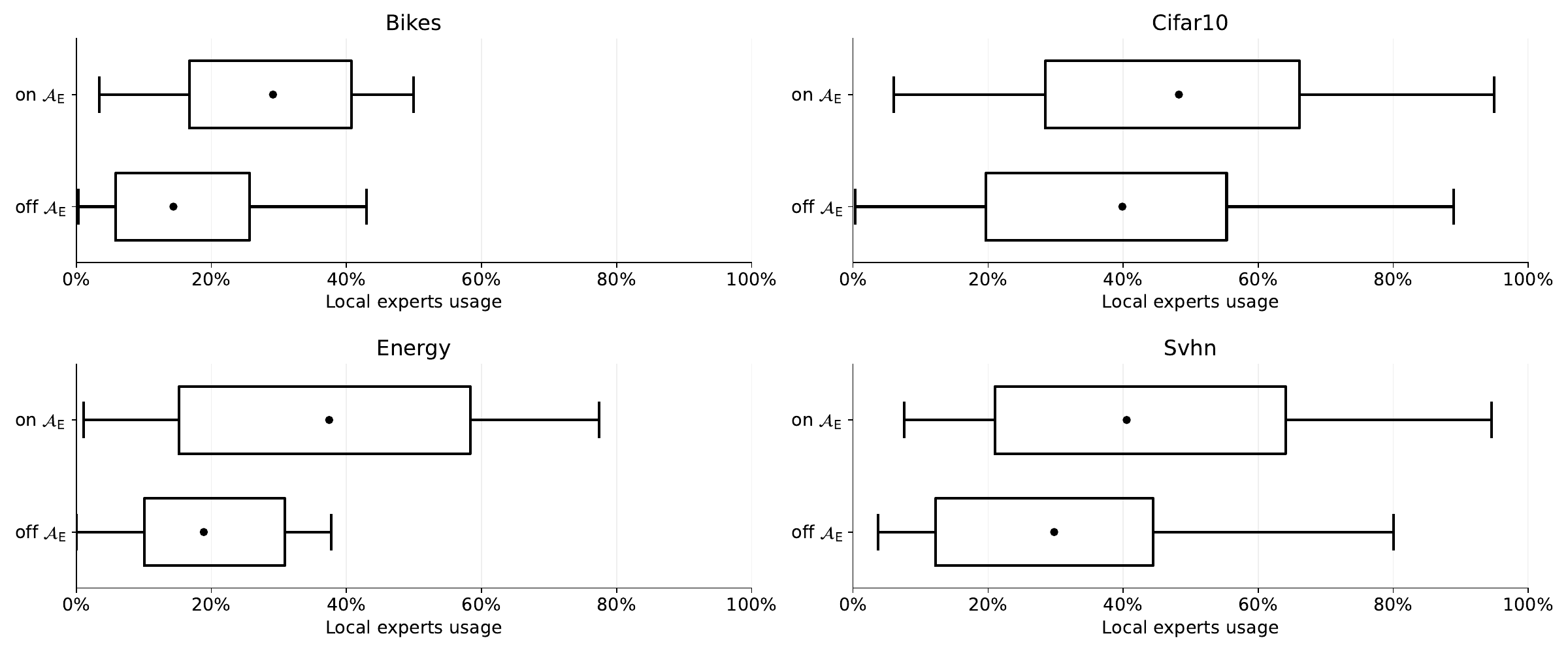}
    \caption{Expert diversity gradient experiment: Absolute local expert usage under \ambiguityon{\pambiguity} and \ambiguityoff{\pambiguity}, computed on validation samples and reported as a percentage.}
    \label{fig:predictor-gradient-local-usage}
\end{figure}

\begin{figure}[H]
    \centering
    \includegraphics[width=0.99\linewidth]{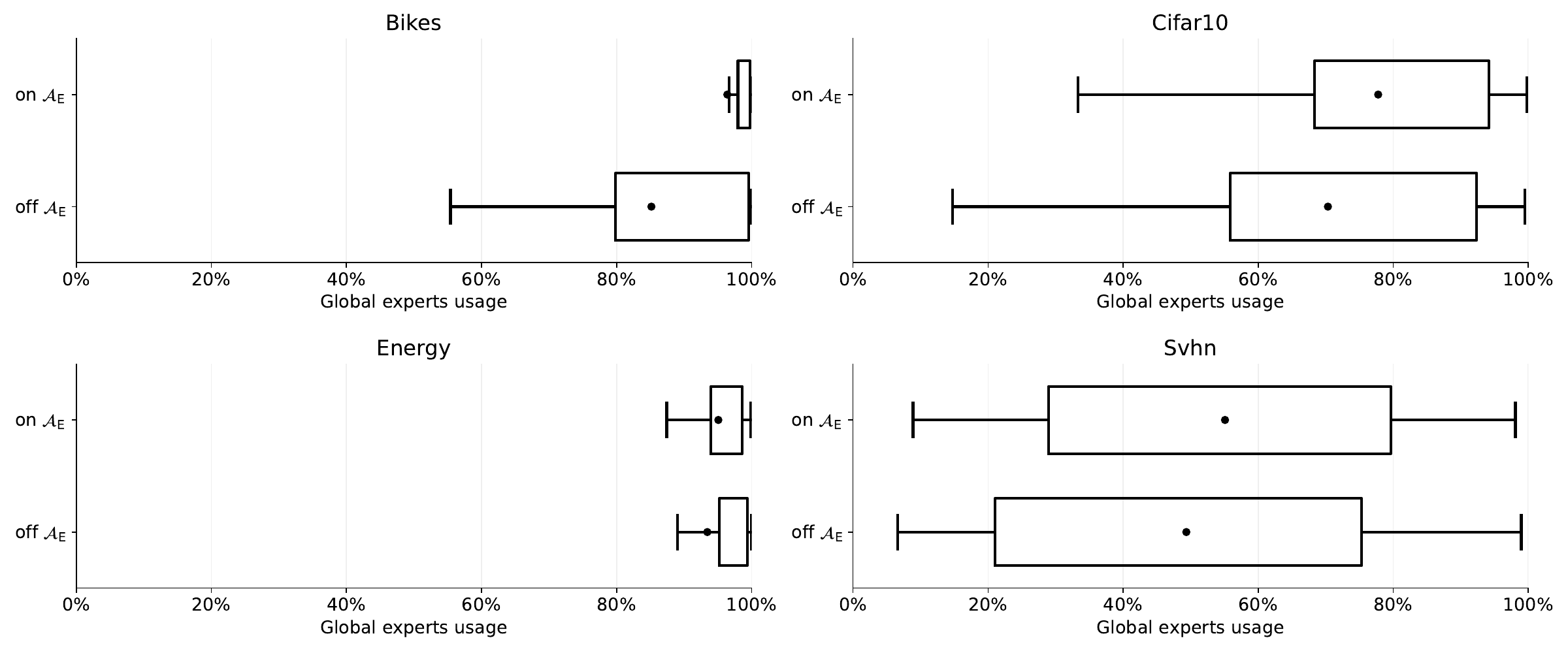}
    \caption{Expert diversity gradient experiment: Absolute global expert usage under \ambiguityon{\pambiguity} and \ambiguityoff{\pambiguity}, computed on validation samples and reported as a percentage.}
    \label{fig:predictor-gradient-global-usage}
\end{figure}

Enabling expert diversity increases local expert usage across the four tasks (Figure~\ref{fig:predictor-gradient-local-usage}). Global usage changes little except on Bikes, where it approaches its maximum (Figure~\ref{fig:predictor-gradient-global-usage}).

\subsection{Router diversity}\label{ap:delegator-gradient-predictors}

\paragraph{Expert decomposition.}

\begin{figure}[H]
    \centering
    \includegraphics[width=0.99\linewidth]{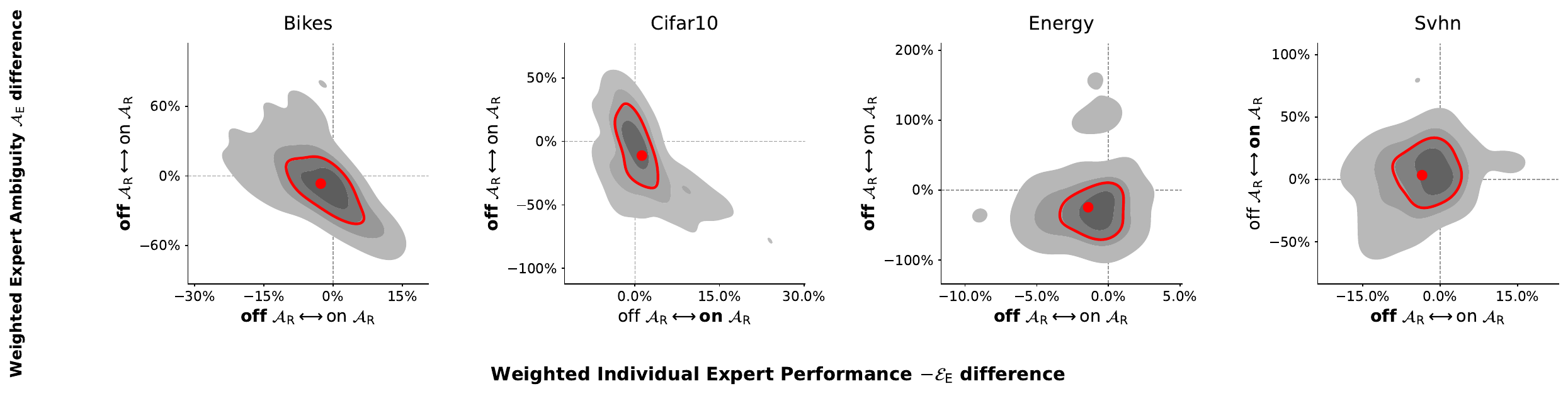}
    \caption{Router diversity gradient experiment: Distribution of the expert performance differences (x-axis) and expert diversity differences (y-axis) between \ambiguityon{\dambiguity} and \ambiguityoff{\dambiguity}, computed on validation samples. The better condition is marked in bold separately on each axis. The red dot denotes the two-dimensional mean, and the red contour encloses the central $50\%$ of the joint distribution.}
    \label{fig:delegator-gradient-predictors}
\end{figure}

Expert-level changes are weak and task-dependent (Figure~\ref{fig:delegator-gradient-predictors}). Energy shows the clearest increase in expert diversity when the router-diversity gradient is blocked; Bikes and CIFAR-10 show smaller or mixed shifts, and SVHN is approximately centered around zero. There is no shared expert-level pattern explaining the final performance.

\paragraph{Expert usage.}\label{ap:usage-delegator}

\begin{figure}[H]
    \centering
    \includegraphics[width=0.99\linewidth]{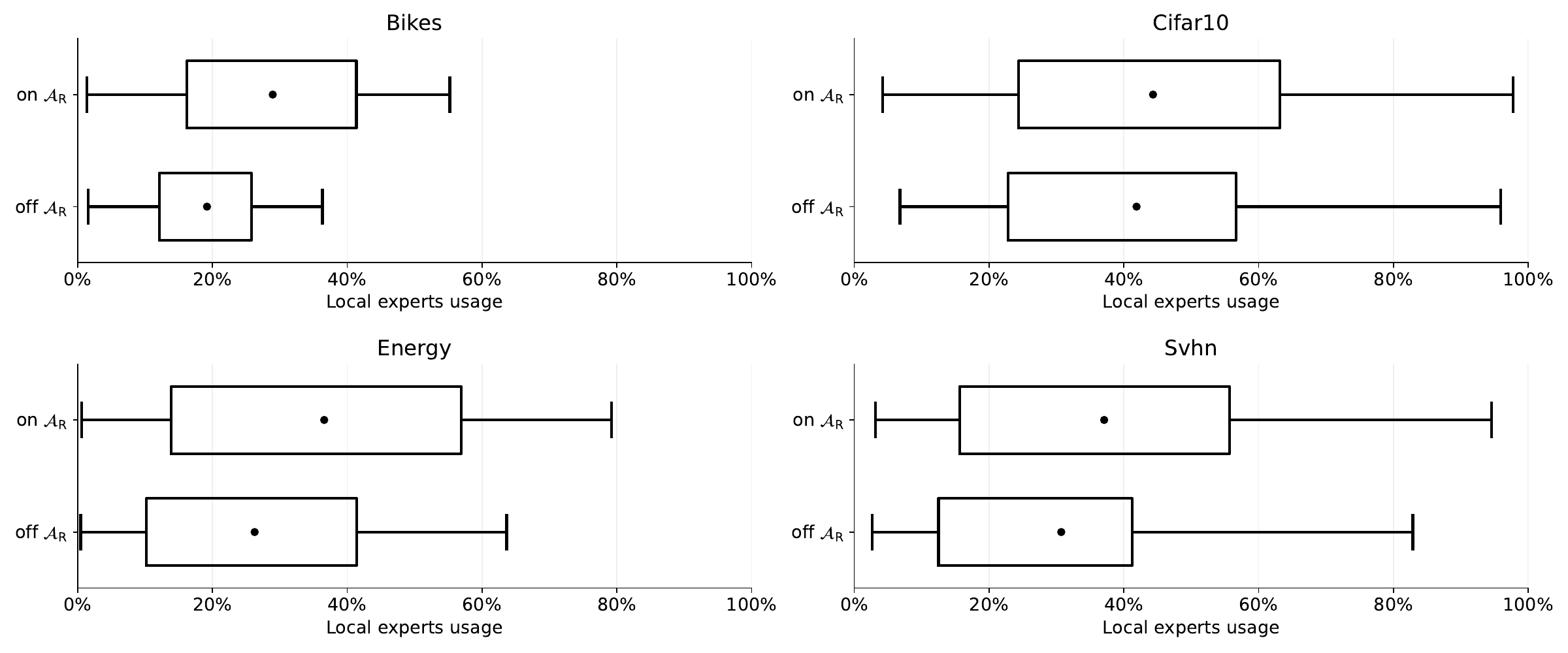}
    \caption{Router diversity gradient experiment: Absolute local expert usage under \ambiguityon{\dambiguity} and \ambiguityoff{\dambiguity}, computed on validation samples and reported as a percentage.} 
    \label{fig:delegator-gradient-local-usage}
\end{figure}

\begin{figure}[H]
    \centering
    \includegraphics[width=0.99\linewidth]{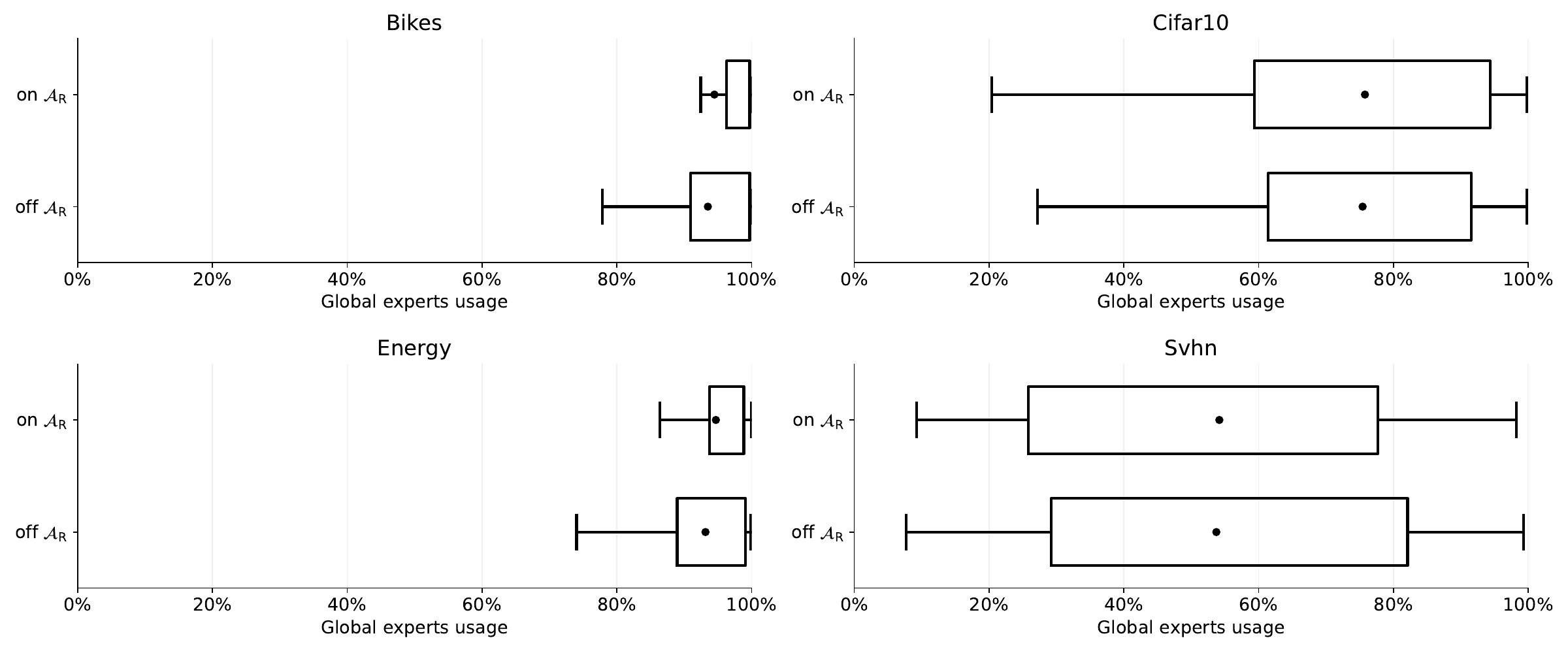}
    \caption{Router diversity gradient experiment: Absolute global expert usage under \ambiguityon{\dambiguity} and \ambiguityoff{\dambiguity}, computed on validation samples and reported as a percentage.}
    \label{fig:delegator-gradient-global-usage} 
\end{figure}

Enabling router diversity has a weak effect on local usage except on Bikes, where the increase is more pronounced (Figure~\ref{fig:delegator-gradient-local-usage}). Global usage changes little across tasks (Figure~\ref{fig:delegator-gradient-global-usage}). 

\subsection{Router aggregation}\label{ap:usage-aggregation}

\paragraph{Expert decomposition.}

\begin{figure}[H]
    \centering
    \includegraphics[width=0.99\linewidth]{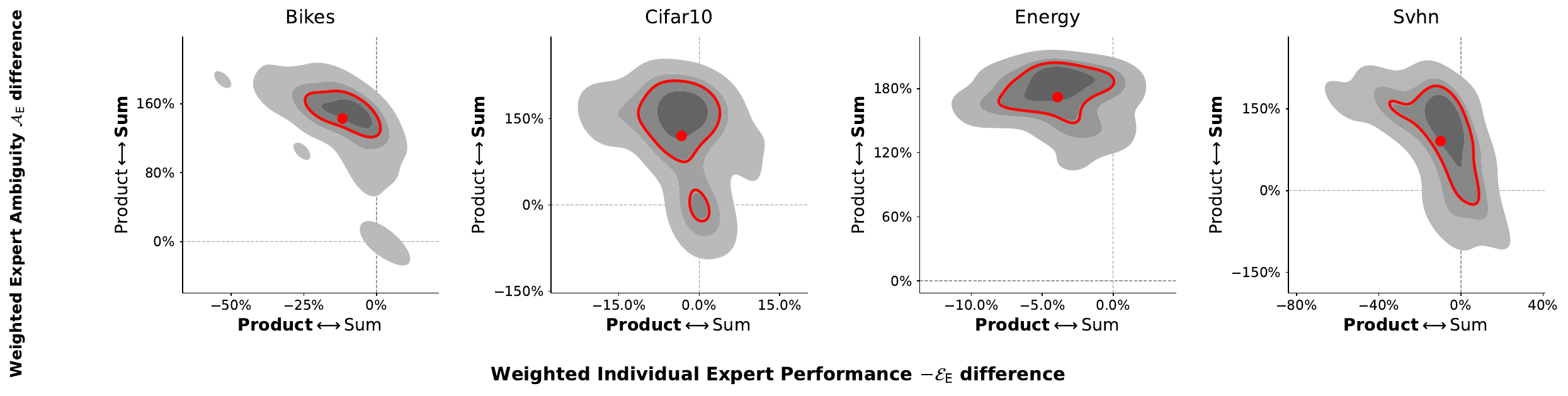}
    \caption{Router aggregation experiment: Distribution of the expert performance differences (x-axis) and expert diversity differences (y-axis) between arithmetic and normalized geometric means, computed on validation samples. The better aggregation rule is marked in bold separately on each axis. The red dot denotes the two-dimensional mean, and the red contour encloses the central $50\%$ of the joint distribution.}
    \label{fig:aggregation-predictors}
\end{figure}

The arithmetic mean produces greater expert diversity but slightly poorer weighted individual expert performance (Figure~\ref{fig:aggregation-predictors}). On Bikes, CIFAR-10, and Energy, the diversity increase offsets this performance cost. SVHN has a smaller diversity difference and more variable individual performance, consistent with its inconclusive total result.

\paragraph{Router diagnostics.}

\begin{figure}[H] 
    \centering
    \includegraphics[width=0.99\linewidth]{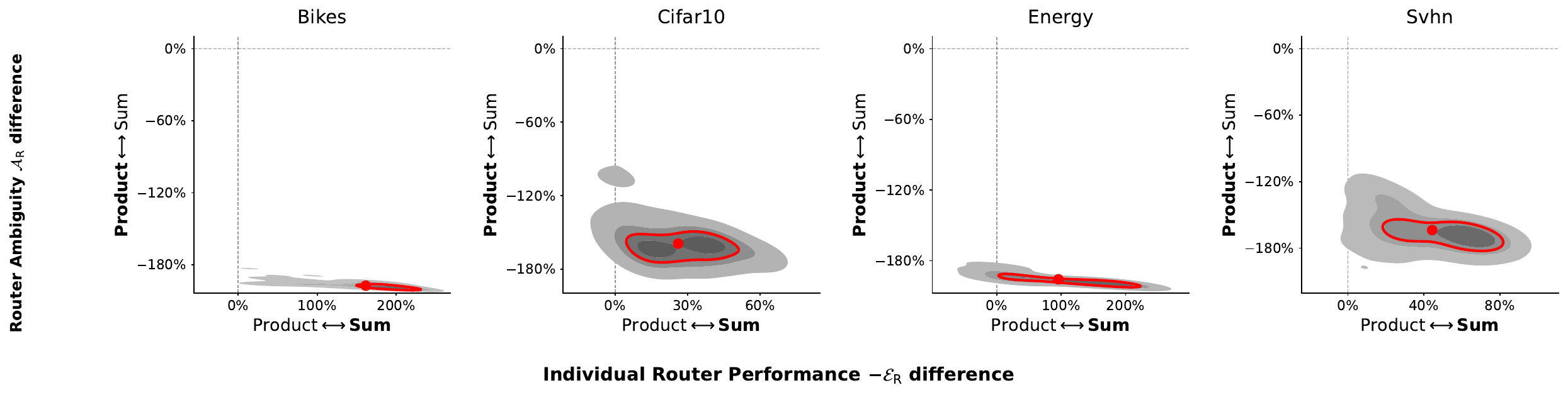}
    \caption{Router aggregation experiment: Distribution of the router performance differences (x-axis) and router diversity differences (y-axis) between arithmetic and normalized geometric means, computed on validation samples. The better aggregation rule is marked in bold separately on each axis. The red dot denotes the two-dimensional mean, and the red contour encloses the central $50\%$ of the joint distribution. Under the geometric mean, these terms do not generally decompose the ensemble loss and should be interpreted only as descriptive diagnostics.}
    \label{fig:aggregation-delegators}
\end{figure}

The arithmetic mean gives better individual router performance and less measured diversity (Figure~\ref{fig:aggregation-delegators}). These quantities form an exact loss decomposition only under the arithmetic mean. Under the geometric mean, $\dperformance-\dambiguity$ need not equal the ensemble loss, so the plotted terms are descriptive diagnostics rather than an exact attribution of that loss. Greater measured diversity therefore cannot be interpreted as a larger subtractive contribution to the final loss.

\paragraph{Expert usage.}

\begin{figure}[H]
    \centering
    \includegraphics[width=0.99\linewidth]{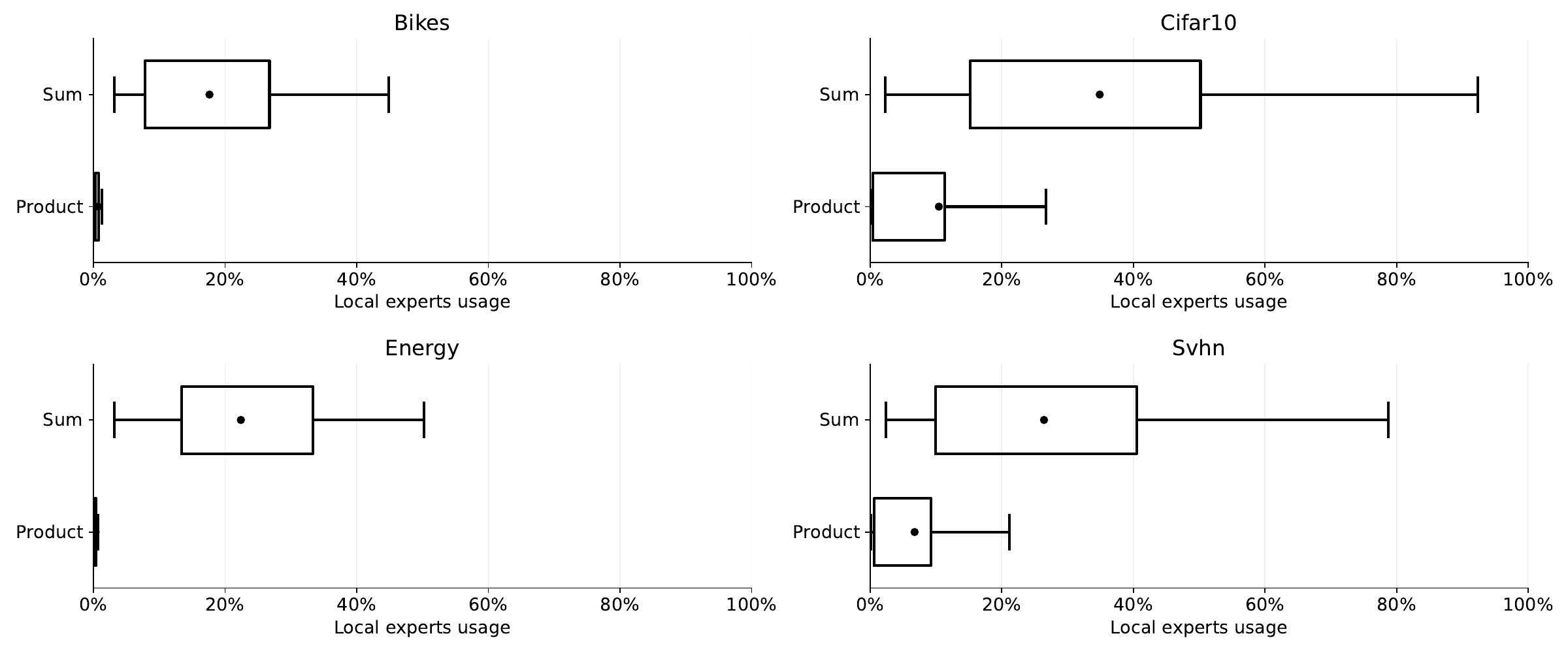}
    \caption{Router aggregation experiment: Absolute local expert usage under arithmetic and normalized geometric means, computed on validation samples and reported as a percentage.}
    \label{fig:aggregation-local-usage}
\end{figure}

\begin{figure}[H]
    \centering
    \includegraphics[width=0.99\linewidth]{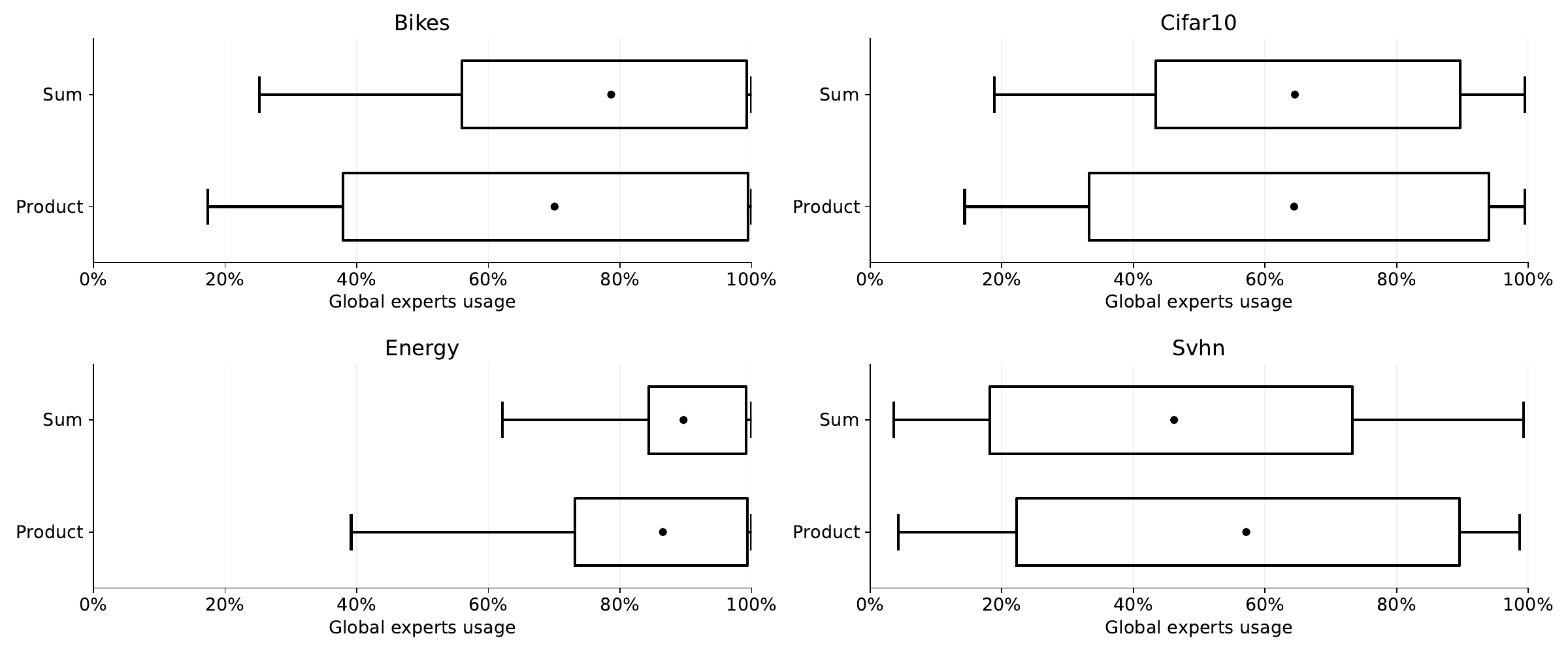}
    \caption{Router aggregation experiment: Absolute global expert usage under arithmetic and normalized geometric means, computed on validation samples and reported as a percentage.}
    \label{fig:aggregation-global-usage}
\end{figure}

The geometric mean concentrates routing weights more strongly than the arithmetic mean, reducing local usage across tasks (Figure~\ref{fig:aggregation-local-usage}). Normalized effective usage is around $1\%$ on the regression tasks and $10\%$ on the image tasks under the geometric mean; these percentages measure weight concentration, not the fraction of experts executed. Global usage remains broadly similar between aggregation rules (Figure~\ref{fig:aggregation-global-usage}), so more concentrated routing per sample need not imply less balanced usage across the dataset.

\section{Gradient routing}\label{ap:gradient-routing}

We enumerate the four combinations of \ambiguityon{\pambiguity} or \ambiguityoff{\pambiguity} and \ambiguityon{\dambiguity} or \ambiguityoff{\dambiguity}. Expert diversity never updates expert parameters; its on/off setting controls only whether it updates routers. We assume the arithmetic mean of router weights and $L\geq 1$, so the ensemble loss satisfies
\begin{equation}
    \ell
    = \pperformance(\bar{\va})-\pambiguity(\bar{\va})
    = \dperformance-\dambiguity,
    \qquad
    \dperformance
    = \frac{1}{L}\sum_{l=1}^L
      \left[\pperformance(\va_l)-\pambiguity(\va_l)\right].
\end{equation}

The operator $\sg_{\vTheta}$ preserves its argument's value but stops its gradient with respect to expert parameters. Similarly, $\sg_{\vPhi}$ stops gradients with respect to router parameters, while $\sg$ stops all gradients. The expressions below define $\ell_{\mathrm{train}}$: each has the same numerical value as $\ell$, but supplies the gradients prescribed by its setting.

All settings also include the auxiliary losses, including the load-balancing penalty in Equation~\ref{eq:training-objective}. These are added unchanged and are not subject to the stop-gradient operations below.

\paragraph{\ambiguityon{\pambiguity}, \ambiguityon{\dambiguity}}
Expert diversity updates only routers:
\begin{equation}
    \ell_{\mathrm{train}}
    = \pperformance(\bar{\va})
      - \sg_{\vTheta}\left[\pambiguity(\bar{\va})\right].
\end{equation}

\paragraph{\ambiguityoff{\pambiguity}, \ambiguityon{\dambiguity}}
Expert diversity contributes no gradient:
\begin{equation}
    \ell_{\mathrm{train}}
    = \pperformance(\bar{\va})
      - \sg\left[\pambiguity(\bar{\va})\right].
\end{equation}

\paragraph{\ambiguityon{\pambiguity}, \ambiguityoff{\dambiguity}}
Experts learn only from their weighted individual performance. Routers learn from their individual performance, which still accounts for expert diversity:
\begin{equation}
\begin{aligned}
    \ell_{\mathrm{train}}
    ={}& \sg_{\vPhi}\left[
           \pperformance(\bar{\va})
           - \sg_{\vTheta}\left[\pambiguity(\bar{\va})\right]
         \right]\\
       &+ \sg_{\vTheta}\left[
           \dperformance
           - \sg_{\vPhi}\left[\dambiguity\right]
           + \sg_{\vPhi}\left[\dambiguity-\dperformance\right]
         \right].
\end{aligned}
\end{equation}

\paragraph{\ambiguityoff{\pambiguity}, \ambiguityoff{\dambiguity}}
Experts learn only from their weighted individual performance. Routers also ignore expert diversity within each individual weight proposal:
\begin{equation}
\begin{aligned}
    \ell_{\mathrm{train}}
    ={}& \sg_{\vPhi}\left[
           \pperformance(\bar{\va})
           - \sg\left[\pambiguity(\bar{\va})\right]
         \right]\\
       &+ \sg_{\vTheta}\left[
           \frac{1}{L}\sum_{l=1}^L
           \left(
               \pperformance(\va_l)
               - \sg_{\vPhi}\left[\pambiguity(\va_l)\right]
           \right)
           - \sg_{\vPhi}\left[\dambiguity\right]
           + \sg_{\vPhi}\left[\dambiguity-\dperformance\right]
         \right].
\end{aligned}
\end{equation}

In the \ambiguityoff{\dambiguity} settings, the second bracket has numerical value zero but supplies the required router gradients. The correction term $\sg_{\vPhi}[\dambiguity-\dperformance]$ therefore preserves the reported ensemble loss without changing either parameter group's updates.

\paragraph{Geometric mean of router weights.}
The aggregation experiment uses \ambiguityoff{\pambiguity} and \ambiguityoff{\dambiguity} with both aggregation rules. For the geometric mean, the expert loss decomposition still holds, but the router identity $\ell=\dperformance-\dambiguity$ generally does not. An equivalent expression for the implemented gradient routing, which does not require that identity, is
\begin{equation}
\begin{aligned}
    Q &= \frac{1}{L}\sum_{l=1}^L\pperformance(\va_l),\\
    \ell_{\mathrm{train}}
    &= \sg_{\vPhi}\!\left[\pperformance(\bar{\va})-\sg[\pambiguity(\bar{\va})]\right]
       +\sg_{\vTheta}\!\left[Q-\sg[Q]\right].
\end{aligned}
\end{equation}
The first term has the value of the ensemble loss and updates only experts through their weighted individual losses. The second term has a value of zero and trains each router through its own weighted expert losses, without expert- or router-diversity gradients. The load-balancing penalty is added unchanged and uses the aggregated routing weights.

\section{Router performance--diversity decomposition under the arithmetic mean}\label{ap:delegator-decomposition}

Fix a sample $(\vx,\vy)$ and $L\geq 1$ routers. Under the arithmetic mean, their weight proposals are averaged:
\begin{equation}
    \bar{\va}=\frac{1}{L}\sum_{l=1}^L\va_l,
    \qquad
    \hat{\vy}^{(l)}=F(\vx;\va_l),
    \qquad 
    \hat{\vy}=F(\vx;\bar{\va}).
\end{equation}
We show that this rule makes the final prediction the appropriate aggregate of the router-induced predictions, and then derive $\ell(\hat{\vy},\vy)=\dperformance-\dambiguity$.

\paragraph{Regression.}
Expert outputs are combined by a weighted arithmetic mean. Writing $a_{l,m}$ for the weight assigned to expert $m$ by router $l$, we obtain
\begin{equation}
\begin{aligned}
    \hat{\vy}
    &=\sum_{m=1}^M\bar a_m\hat{\vy}_m
    =\frac{1}{L}\sum_{l=1}^L\sum_{m=1}^M a_{l,m}\hat{\vy}_m
    =\frac{1}{L}\sum_{l=1}^L\hat{\vy}^{(l)} \\
    \mathbf{0} &= \frac{1}{L}\sum_{l=1}^L\hat{\vy}^{(l)} - \hat{\vy} = \sum_{l=1}^L(\hat{\vy}^{(l)} - \hat{\vy})
\end{aligned}
\end{equation}
Expanding each individual squared error around $\hat{\vy}$ gives
\begin{equation}
\begin{aligned}
    \dperformance
    &=\frac{1}{Ld_y}\sum_{l=1}^L
      \left\|(\hat{\vy}^{(l)}-\hat{\vy})+(\hat{\vy}-\vy)\right\|_2^2\\
    &= \frac{1}{Ld_y}\sum_{l=1}^L \left\| \hat{\vy}^{(l)}-\hat{\vy} \right\|_2^2  + \frac{2}{Ld_y} (\hat{\vy}-\vy)^\top  \underbrace{\sum_{l=1}^L (\hat{\vy}^{(l)}-\hat{\vy})}_{\mathbf{0}}  + \frac{1}{Ld_y}\sum_{l=1}^L \left\| \hat{\vy}-\vy \right\|_2^2 \\
    &= \dambiguity + SE(\hat{\vy}, \vy)
\end{aligned}
\end{equation}
The cross term vanishes, yielding
\begin{equation}\label{eq:delegator-se-derivation}
    \mathrm{SE}(\hat{\vy},\vy)=\dperformance-\dambiguity.
\end{equation}

\paragraph{Classification.}
Expert outputs are combined by a normalized weighted geometric mean. Assume positive expert probabilities, so the logarithms below are defined. For class $c$, the prediction induced by router $l$ is
\begin{equation}
    \hat y_c^{(l)}
    =\frac{1}{Z_l}\prod_{m=1}^M\hat y_{m,c}^{a_{l,m}},
    \qquad
    Z_l=\sum_{c=1}^{d_y}\prod_{m=1}^M\hat y_{m,c}^{a_{l,m}}.
\end{equation}
Taking the geometric mean across routers gives
\begin{equation}
    \prod_{l=1}^L\big(\hat y_c^{(l)}\big)^{1/L}
    =\left(\prod_{l=1}^L Z_l^{-1/L}\right)
      \prod_{m=1}^M\hat y_{m,c}^{\frac{1}{L}\sum_{l=1}^L a_{l,m}}
    =\left(\prod_{l=1}^L Z_l^{-1/L}\right)
      \prod_{m=1}^M\hat y_{m,c}^{\bar a_m}.
\end{equation}
The factor involving $Z_l$ is the same for every class and cancels when the probabilities are normalized. Hence the final prediction is the normalized geometric mean of the induced predictions:
\begin{equation}
    \hat y_c=\frac{1}{Z_{\mathrm{R}}}
      \prod_{l=1}^L\big(\hat y_c^{(l)}\big)^{1/L},
    \qquad
    Z_{\mathrm{R}}=\sum_{c=1}^{d_y}
      \prod_{l=1}^L\big(\hat y_c^{(l)}\big)^{1/L}.
\end{equation}
Taking logarithms yields
\begin{equation}
    \frac{1}{L}\sum_{l=1}^L\log\hat y_c^{(l)}
    =\log\hat y_c+\log Z_{\mathrm{R}}.
\end{equation}
Using $\sum_c y_c=1$, the mean individual router loss is therefore
\begin{equation}
\begin{aligned}
    \dperformance
    &=-\sum_{c=1}^{d_y}y_c
       \left(\frac{1}{L}\sum_{l=1}^L\log\hat y_c^{(l)}\right)\\
    &=\mathrm{CE}(\hat{\vy},\vy)-\log Z_{\mathrm{R}}.
\end{aligned}
\end{equation}
Similarly, using $\sum_c\hat y_c=1$, the mean router diversity is
\begin{equation}
\begin{aligned}
    \dambiguity
    &=\frac{1}{L}\sum_{l=1}^L
      \mathrm{KL}(\hat{\vy}\,\|\,\hat{\vy}^{(l)})\\
    &=\sum_{c=1}^{d_y}\hat y_c
      \left(\log\hat y_c-\frac{1}{L}\sum_{l=1}^L\log\hat y_c^{(l)}\right)
      =-\log Z_{\mathrm{R}}.
\end{aligned}
\end{equation}
Subtracting the two expressions gives
\begin{equation}\label{eq:delegator-ce-derivation}
    \mathrm{CE}(\hat{\vy},\vy)=\dperformance-\dambiguity.
\end{equation}
Equations~\ref{eq:delegator-se-derivation} and~\ref{eq:delegator-ce-derivation} establish Equation~\ref{eq:delegator-ambiguity}. Both rely on the arithmetic mean of routing weights; the geometric mean of routing weights does not generally produce the required mean of the induced predictions.

%
%
%

\end{document}